\documentclass[letterpaper]{article} 
\usepackage{aaai2027}  
\nocopyright
\usepackage[hyphens]{url}  
\usepackage{graphicx} 
\usepackage{natbib}  
\usepackage{caption} 
\usepackage{algorithm}
\usepackage{algorithmic}

\usepackage{amsmath}   
\usepackage{amssymb}   
\usepackage{bm}
\usepackage{multirow}
\usepackage{newfloat}
\usepackage{listings}
\DeclareCaptionStyle{ruled}{labelfont=normalfont,labelsep=colon,strut=off} 
\floatstyle{ruled}
\newfloat{listing}{tb}{lst}{}
\floatname{listing}{Listing}

\usepackage{booktabs}

\usepackage[hyphens]{url}
\usepackage{graphicx}
\usepackage{natbib}
\usepackage{caption}

\usepackage{tikz}
\usetikzlibrary{arrows.meta,positioning,calc}

\usepackage{fvextra}

\DefineVerbatimEnvironment{PromptBlock}{Verbatim}{
  fontsize=\footnotesize,
  breaklines=true,
  breakanywhere=true,
  breaksymbolleft={},
  breaksymbolright={},
  frame=single,
  framesep=2mm,
  baselinestretch=0.95
}
\usepackage{}
\usepackage{algorithm}
\usepackage{algorithmic}
\usepackage{amsmath}
\usepackage{amssymb}
\usepackage{bm}
\usepackage{booktabs}
\usepackage{multirow}
\usepackage{float}
\newcommand{\src}{\mathrm{src}}
\newcommand{\tgt}{\mathrm{tgt}}

\newcommand{\eps}{\varepsilon}
\newcommand{\placeholdergraphic}[2]{%
  \IfFileExists{#1}{\includegraphics[width=\textwidth]{#1}}{%
    \fbox{\parbox[c][1.55in][c]{0.965\textwidth}{\centering\textbf{#2}\\[0.45em]
    Replace this box by saving a full-width PDF at\\\texttt{\detokenize{#1}}.}}}}
\newif\ifshowslots
\ifdefined\PCEDITHIDESLOTS\showslotsfalse\else\showslotstrue\fi
\usepackage{xcolor}
\title{PC-Edit: Prompt-Contrastive Region Discovery and Region-Guided Editing}

\author{
Jian Zhang\textsuperscript{\rm 1},
Zhijun Zhang\textsuperscript{\rm 1}
}

\affiliations{
\textsuperscript{\rm 1}School of Electronic and Information Engineering, 
South China University of Technology, Guangzhou, China\\
202411084134@mail.scut.edu.cn,
auzjzhang@scut.edu.cn\\
Project page: \textcolor{blue}{https://jianzhang-chick.github.io/PC-Edit/}
}

\begin{document}

\maketitle

\begin{abstract}
Replacing an object with one that differs in category or shape requires complete source removal, natural target formation unconstrained by the source silhouette, and preservation of unrelated content. Existing training-free editors either localize edits from terminal predictions under source and target prompts or preserve unrelated content through spatially unselective source-feature reuse without explicit region discovery. Before reaching the terminal predictions, prompt-induced semantic differences undergo additional network transformations that may obscure their spatial localization, reducing localization precision. Spatially unselective feature reuse forces a trade-off between edit completeness and background preservation. Therefore, we propose PC-Edit, a prompt-contrastive framework for training-free MM-DiT editing. PC-Edit contrasts the image-token attention outputs under the source and target prompts, capturing prompt-induced semantic differences directly where text-conditioned information is delivered to image tokens. The same contrast identifies a source-erasure region during inversion and a target-emergence region during denoising. Their union suppresses source remnants while allowing the target object to form naturally. PC-Edit further couples region discovery and background preservation within each sampling step by estimating the current edit region from preceding attention blocks and immediately injecting cached source K/V features outside it in subsequent blocks, thereby protecting unrelated content before the latent update. Experiments on PIE-Bench and our EditRegion-Bench, with human-verified edit-region annotations for single- and multi-object addition and replacement, show that PC-Edit achieves the best editing quality and background preservation among methods without user-specified edit regions.
\end{abstract}

\section{Introduction}
\label{sec:intro}

Text-to-image generation has rapidly shifted from UNet-based latent diffusion~\citep{rombach2022ldm} to transformer backbones~\citep{peebles2023dit} trained with flow matching~\citep{lipman2023flow,liu2023rectified}, such as SD3~\citep{esser2024sd3} and FLUX~\citep{flux2024}. Building on these models, training-free text-guided editing enables diverse local and global modifications through improved inversion~\citep{mokady2023nulltext,ju2024pnpinversion,rout2025rfinversion,wang2025taming} and internal-feature manipulation~\citep{hertz2023prompt,cao2023masactrl,tumanyan2023pnp,avrahami2025stableflow}. Yet object replacement remains difficult when the source and target objects differ substantially in category, silhouette, and spatial extent. Successful replacement must fully remove the source object, allow the target to form in its own shape, and preserve unrelated content.

Without user-provided edit region masks, the central challenge of replacement tasks is automatic region discovery. Mask-conditioned methods such as FLUX.1-Fill~\citep{flux2024} and KV-Edit~\citep{zhu2025kvedit} can preserve the background well, but shift the burden of localization to the user and constrain the target object to a prescribed region. Automatic methods instead derive edit regions from prompt-conditioned terminal predictions. DiffEdit~\citep{couairon2023diffedit} compares noise predictions under two prompts, while Follow-Your-Shape (FYS)~\citep{zhang2025fys} compares token-wise velocity differences across the source inversion and target editing trajectories. However, by the time terminal predictions are produced, subsequent network transformations may have obscured the spatial response of prompt-induced semantic differences. The cross-trajectory comparison in FYS additionally changes the latent state, introducing interference from trajectory mismatch and solver error. These factors reduce spatial localization precision.

\begin{figure*}[t]
    \centering
    \scalebox{1}[0.94]{\includegraphics[width=\textwidth]{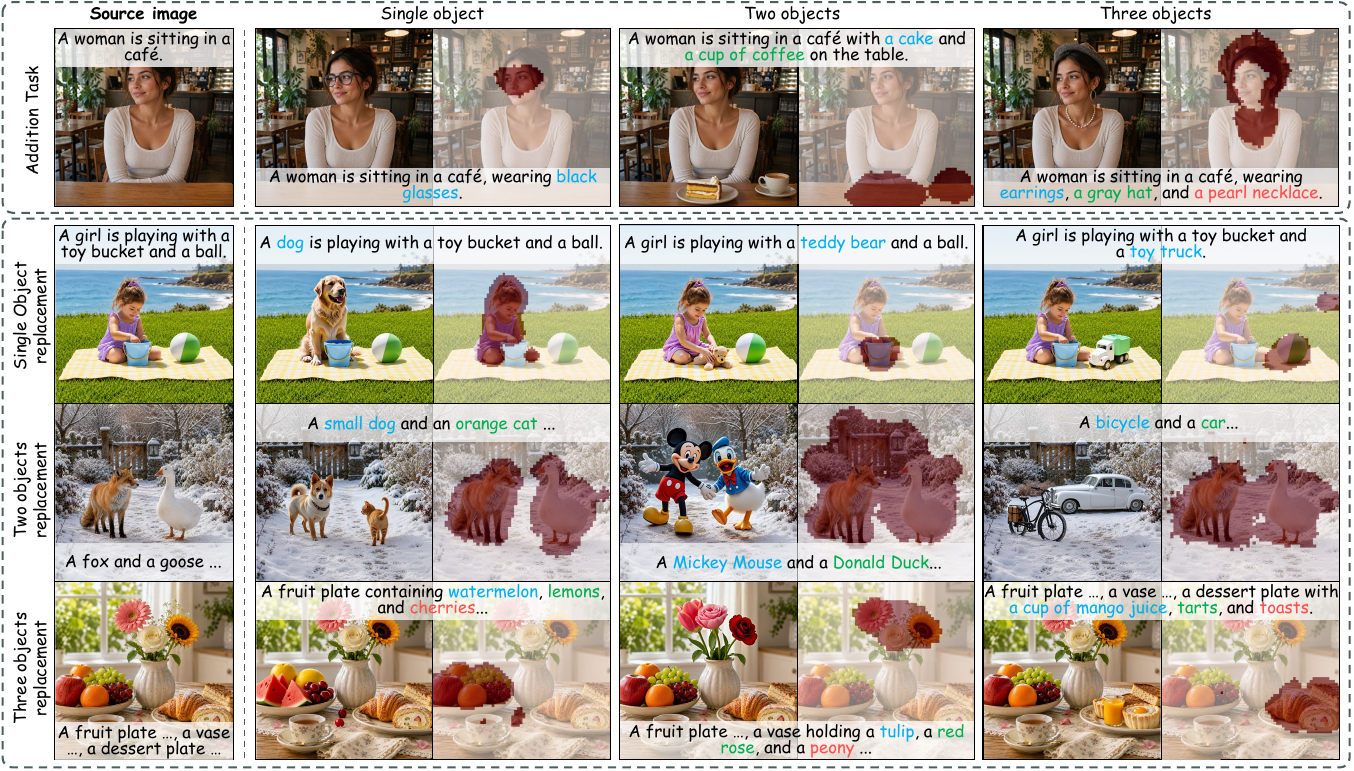}}
    \caption{
    \textbf{Visual results of PC-Edit.} Each example pairs an edited image (left) with its automatically discovered edit region (right), covering single- and multi-object addition and replacement.
    }
\label{fig:teaser}
\end{figure*}

To address these issues, we propose PC-Edit, a prompt-contrastive framework for training-free MM-DiT editing. As illustrated in Fig.~\ref{fig:teaser}, PC-Edit supports single- and multi-object addition and replacement without user-provided masks while automatically discovering edit regions. At the same latent state and timestep, PC-Edit contrasts the image-token attention outputs under the source and target prompts. Since these outputs carry text-conditioned information written into image tokens through joint attention, their difference directly reveals which image locations respond to the prompt change, providing a more direct and accurate localization signal than terminal predictions. However, accurate localization alone is insufficient for large-shape replacement because the source-erasure and target-emergence footprints need not coincide. A mask that only follows the emerging target can leave source-object remnants, whereas a mask restricted to the source footprint can suppress the target’s natural shape. PC-Edit applies the same contrast during inversion to identify a source-erasure region and during denoising to track a target-emergence region. Their union suppresses source-object remnants while allowing the target to extend beyond the source silhouette. 

PC-Edit further couples region discovery and preservation within each sampling step. It estimates the current edit region from preceding attention blocks and immediately injects cached source K/V features outside it in subsequent blocks. This same-step design uses the latest region estimate to guide each latent update, thereby reducing background drift. To evaluate automatic region discovery and editing quality, we introduce EditRegion-Bench, a 484-case benchmark for single- and multi-object addition
and replacement, with manually annotated edit regions. Extensive experiments on EditRegion-Bench and PIE-Bench validate the effectiveness of the proposed localization and editing mechanisms.

Our main contributions are as follows:
\begin{itemize}

\item We identify differences between MM-DiT image-token attention outputs under source and target prompts as an effective edit-region localization signal and develop a corresponding prompt-contrastive localization method.

\item We develop an editing mechanism that separately models source-erasure and target-emergence regions and injects source K/V features outside their union within the same sampling step, allowing the target to form without being constrained by the source silhouette while suppressing source remnants and preserving unrelated content.

\item We introduce EditRegion-Bench, a 484-case benchmark with manually annotated edit regions for single- and multi-object editing, enabling evaluation of edit-region localization, background fidelity, and visual quality.

\item Experiments on EditRegion-Bench and PIE-Bench demonstrate that PC-Edit achieves the best editing quality and background preservation among methods without user-specified edit regions.

\end{itemize}

\begin{figure*}
    \centering
    \scalebox{1}[0.9]{\includegraphics[width=1\linewidth]{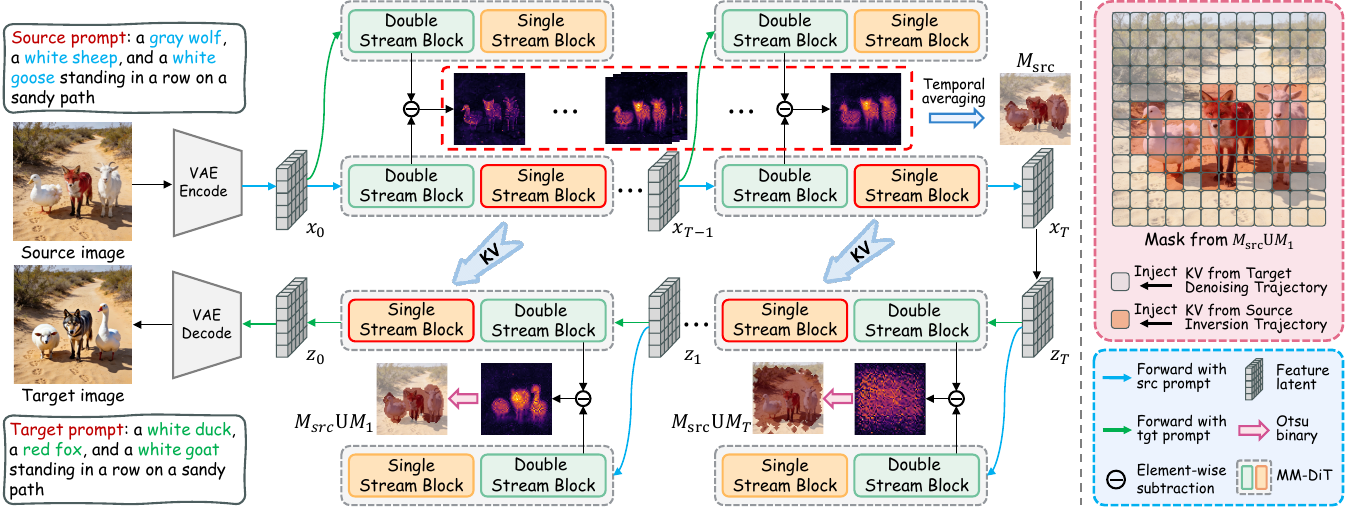}}
    \caption{
    \textbf{Overview of PC-Edit.} Inversion caches source K/V features and estimates the source-erasure mask $M_{\mathrm{src}}$. Denoising tracks the target-emergence mask, combines it with $M_{\mathrm{src}}$, and injects source K/V outside the resulting region within the same sampling step to preserve unrelated content.
    }
    \label{fig:overview}
\end{figure*}

\section{Related Work}
\label{sec:related}

\paragraph{Training-free image editing.}
Training-free editing modifies images using a frozen pretrained text-to-image model, typically through source-image inversion and internal-feature manipulation. On UNet-based diffusion models, Prompt-to-Prompt~\citep{hertz2023prompt} controls cross-attention, Null-text Inversion~\citep{mokady2023nulltext} and Direct Inversion~\citep{ju2024pnpinversion} improve reconstruction, and MasaCtrl~\citep{cao2023masactrl} and Plug-and-Play~\citep{tumanyan2023pnp} reuse internal features to preserve structure. Recent methods extend these ideas to diffusion transformers and rectified-flow models. DiT4Edit~\citep{feng2025dit4edit}, RF-Inversion~\citep{rout2025rfinversion}, and RF-Edit~\citep{wang2025taming} improve inversion or trajectory recovery, while FlowEdit~\citep{kulikov2025flowedit} constructs a direct source-to-target transport path. Stable Flow~\citep{avrahami2025stableflow} preserves source structure through feature injection. However, these methods focus on improving inversion fidelity, source-to-target transport, or content preservation through attention and feature control, without explicit edit-region localization.

\paragraph{Automatic region discovery and region-guided editing.}
Mask-conditioned methods such as FLUX.1-Fill~\citep{flux2024} and KV-Edit~\citep{zhu2025kvedit} achieve strong background preservation, but require a user-provided region and constrain the target object to that region. Mask-free methods instead infer the edit region from prompt-conditioned predictions. DiffEdit~\citep{couairon2023diffedit} compares noise predictions conditioned on the source and target prompts, while FYS~\citep{zhang2025fys} compares velocity predictions along source inversion and target denoising trajectories. However, because these signals are read from terminal predictions, subsequent network transformations may have obscured the spatial response of prompt-induced semantic differences. The cross-trajectory comparison in FYS also changes the latent state, which may introduce trajectory mismatch and solver error. Moreover, a single edit region does not explicitly distinguish the source-erasure region from the target-emergence region. PC-Edit instead reads prompt contrast from image-token attention outputs~\citep{vaswani2017attention} at the same latent state and timestep, models source erasure and target emergence separately, and applies region-guided source K/V injection within the same sampling step.

\section{Methodology}
\label{sec:method}

Given a source image $I_{\mathrm{src}}$ and source and target prompts $(\bm{c}_{\mathrm{src}}, \bm{c}_{\mathrm{tgt}})$, PC-Edit performs the requested object addition or replacement while preserving unrelated content. As illustrated in Fig.~\ref{fig:overview}, the method consists of source inversion and target denoising. During inversion, we map the source latent $\bm{x}_0$ to $\bm{x}_T$, cache source K/V features, and estimate a source-erasure region. Starting from $\bm{z}_T=\bm{x}_T$, denoising dynamically tracks the target-emergence region and uses the resulting edit region to guide same-step source K/V injection.

\subsection{Motivation}
\label{sec:motivation}

\paragraph{Disentangling prompt contrast from state mismatch.}
The trajectory-divergence method~\citep{zhang2025fys} compares the source inversion trajectory $\{\bm{x}_t\}$ under $\bm{c}_{\mathrm{src}}$ with the target denoising trajectory $\{\bm{z}_t\}$ under $\bm{c}_{\mathrm{tgt}}$. Its velocity difference changes both the prompt and the latent state and can be decomposed as
\begin{equation}
\begin{aligned}
    \bm{v}_\theta(\bm{z}_t,\bm{c}_{\mathrm{tgt}})
    - \bm{v}_\theta(\bm{x}_t,\bm{c}_{\mathrm{src}})
    =\;&
    \underbrace{
    \bm{v}_\theta(\bm{x}_t,\bm{c}_{\mathrm{tgt}})
    - \bm{v}_\theta(\bm{x}_t,\bm{c}_{\mathrm{src}})
    }_{\text{prompt contrast, shared state}} \\
    +\;&
    \underbrace{
    \bm{v}_\theta(\bm{z}_t,\bm{c}_{\mathrm{tgt}})
    - \bm{v}_\theta(\bm{x}_t,\bm{c}_{\mathrm{tgt}})
    }_{\text{state mismatch, shared prompt}} .
\end{aligned}
\label{eq:decompose}
\end{equation}
The first term measures the prompt-conditioned change at a shared latent state, whereas the second reflects differences between the inversion and denoising states, including trajectory mismatch and solver error. To avoid this state-dependent interference, PC-Edit varies only the text condition while holding the latent state and timestep fixed.

\paragraph{Reading prompt contrast from attention outputs.}
State matching determines how the source and target prompts should be compared, but still leaves the question of where their difference should be obtained. Existing prediction-difference methods read prompt contrast from terminal noise or velocity predictions~\citep{couairon2023diffedit,zhang2025fys}. Before reaching these predictions, the prompt-conditioned response produced by text--image interaction undergoes further network processing. These later transformations may dilute the prompt-induced semantic contrast and make it less spatially localized. Therefore, we seek an internal readout closer to where text conditioning modulates image tokens. In MM-DiT backbones, text and image tokens interact through joint attention. For each image token, joint attention incorporates text-derived value content into an image-side attention output $\mathrm{AttnOut}$, which carries the text-conditioned modulation of the image representation. Therefore, its source--target difference provides a more direct signal for locating the spatial response induced by a prompt change.

\paragraph{Separating source erasure from target emergence.}
Prediction-difference methods such as DiffEdit~\citep{couairon2023diffedit}
and FYS~\citep{zhang2025fys} derive an edit region, but do not explicitly distinguish the source footprint that must be erased from the region in which the target object should emerge. This distinction becomes critical when the source and target differ substantially in shape or spatial extent. Meanwhile, feature-reuse methods such as Plug-and-Play~\citep{tumanyan2023pnp}, MasaCtrl~\citep{cao2023masactrl}, and Stable Flow~\citep{avrahami2025stableflow} do not explicitly distinguish regions that should retain source features from those that should remain target-driven. Reusing source features in regions that must change can leave source remnants or hinder target formation, whereas weakening reuse reduces the protection of unrelated content. PC-Edit addresses this preservation--editability trade-off by separately modeling source-erasure and target-emergence regions and injecting source K/V features only outside their union.

\subsection{PC-Edit}
\label{sec:pcedit}

We introduce PC-Edit, a prompt-contrastive framework that automatically discovers source-erasure and target-emergence regions during inversion and denoising, respectively, and injects source K/V features outside their union within each region-guided sampling step.

\paragraph{Prompt-contrastive localization signal.}
Let $\mathcal{B}_{\mathrm{loc}}$ denote the set of attention blocks used for region localization. At timestep $t$, PC-Edit runs two forward passes on the same latent, conditioned on $\bm{c}_{\mathrm{src}}$ and $\bm{c}_{\mathrm{tgt}}$, respectively, and computes
\begin{equation}
    \Delta\mathrm{AttnOut}_t
    =
    \frac{1}{|\mathcal{B}_{\mathrm{loc}}|}
    \sum_{b\in\mathcal{B}_{\mathrm{loc}}}
    \left\|
    \mathrm{AttnOut}^{\mathrm{tgt}}_{t,b}
    -
    \mathrm{AttnOut}^{\mathrm{src}}_{t,b}
    \right\|,
\label{eq:delta}
\end{equation}
where the norm over feature channels produces a token-wise heatmap. The latent is $\bm{x}_t$ during inversion and $\bm{z}_t$ during denoising, so the two forward passes differ only in their text conditions.

\paragraph{Source-erasure footprint during inversion.}
Along the source inversion trajectory, we cache the key and value features of the injection blocks $\mathcal{B}_{\mathrm{inj}}$ at every timestep, denoted by $\{K^{\mathrm{src}}_{t,b},V^{\mathrm{src}}_{t,b}\}_{b\in \mathcal{B}_{\mathrm{inj}}}$. These cached features are later reused to anchor regions that should remain unchanged.

For object replacement, we estimate the source-erasure footprint during inversion. Specifically, at each selected timestep $t$, we perform an additional target-conditioned forward pass at the same inversion latent $\bm{x}_t$ to compute $\Delta\mathrm{AttnOut}_t$. Responses near the clean endpoint tend to be weak, whereas those near the noise endpoint become spatially diffuse. Since intermediate states retain recognizable source structure while providing clear source--target contrast, we select an intermediate timestep window $\mathcal{T}_{\mathrm{probe}}$, average the resulting heatmaps, and apply Otsu's threshold~\citep{otsu1975threshold}:
\begin{align}
H_{\mathrm{src}}
&=
\frac{1}{|\mathcal{T}_{\mathrm{probe}}|}
\sum_{t\in\mathcal{T}_{\mathrm{probe}}}
\Delta\mathrm{AttnOut}_t,
\nonumber\\
M_{\mathrm{src}}
&=
\operatorname{Binarize}\!\left(
H_{\mathrm{src}},
\operatorname{Otsu}(H_{\mathrm{src}})
\right).
\label{eq:msrc}
\end{align}
Here, $\operatorname{Binarize}(H,\tau)$ applies an element-wise threshold at $\tau$, assigning one to responses above the threshold and zero otherwise. $M_{\mathrm{src}}$ identifies where source K/V injection must be disabled to remove the original object. For object addition, no source object needs to be erased, so we set $M_{\mathrm{src}}=\mathbf{0}$.

\begin{figure*}[t]
    \centering
    \scalebox{1}[0.9]{\includegraphics[width=\textwidth]{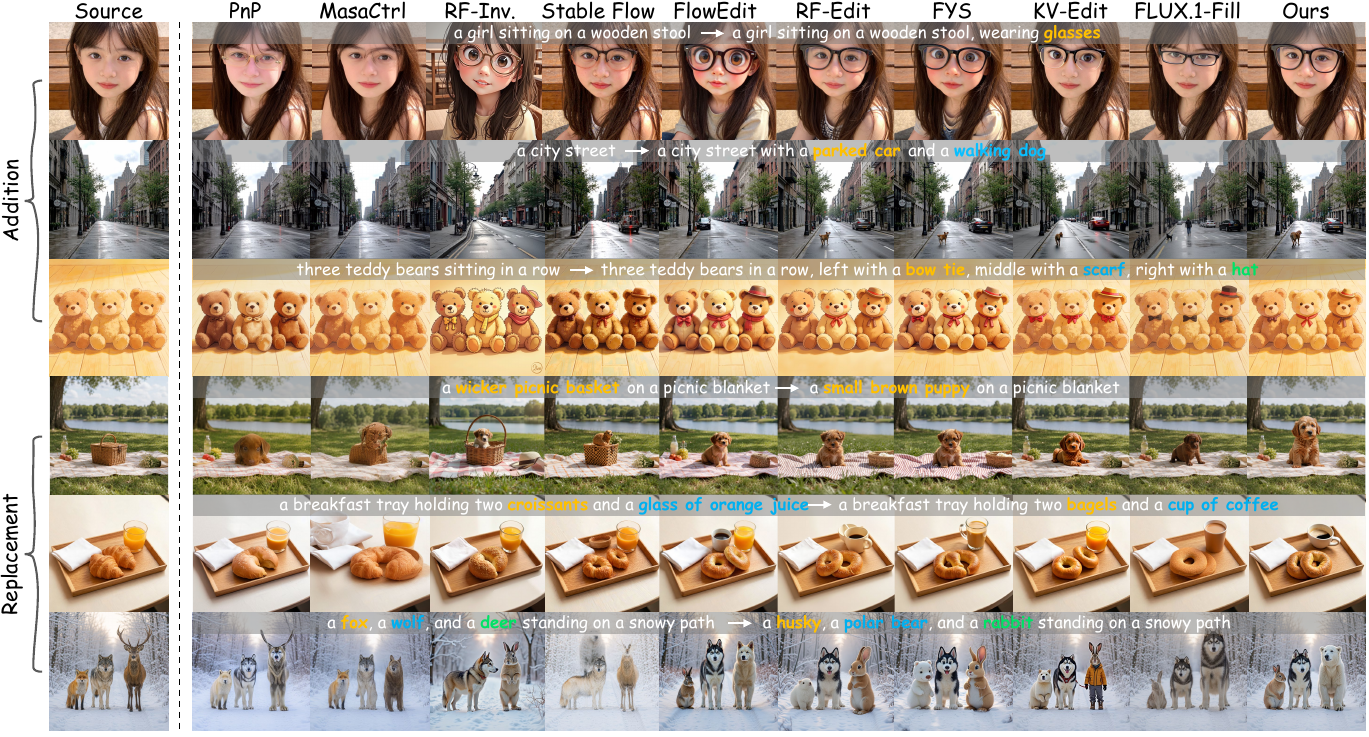}}
    \caption{
    \textbf{Qualitative comparison on addition and replacement.}
    PC-Edit completes single- and multi-object edits while better preserving unrelated content.
    }
    \label{fig:qualitative_comparison}
\end{figure*}

\paragraph{Target-emergence tracking during denoising.}
At denoising step $t$, before source K/V injection, we run two forward passes from the current latent $\bm{z}_t$, conditioned on $\bm{c}_{\mathrm{src}}$ and $\bm{c}_{\mathrm{tgt}}$, respectively, and compute $\Delta\mathrm{AttnOut}_t$ from the localization blocks $\mathcal{B}_{\mathrm{loc}}$. We then apply Otsu’s threshold to obtain the target-emergence mask $M_t$.

Early in denoising, $M_t$ is typically broad, allowing the target object to extend beyond the source silhouette. As denoising proceeds, $M_t$ gradually contracts around the emerging shape of the target object. We combine source erasure and target emergence as
\begin{equation}
    R_t =
    \begin{cases}
        M_{\mathrm{src}} \cup M_t, & t \leq t_s, \\
        R_{t_s}, & t > t_s .
    \end{cases}
\label{eq:region}
\end{equation}
The union covers both the disappearing source object and the emerging target object. Because late-stage contrastive responses become less reliable for localization, we freeze the region at step $t_s$.

\paragraph{Same-step region-guided K/V injection.}
PC-Edit first constructs $R_t$ from the prompt-contrastive responses of the localization blocks $\mathcal{B}_{\mathrm{loc}}$, and then immediately uses it to guide source K/V injection in the subsequent blocks $\mathcal{B}_{\mathrm{inj}}$. For each
$b\in\mathcal{B}_{\mathrm{inj}}$, we compute
\begin{equation}
\begin{aligned}
    K_{t,b} &=
    R_t \odot K^{\mathrm{tgt}}_{t,b}
    +
    (1-R_t) \odot K^{\mathrm{src}}_{t,b}, \\
    V_{t,b} &=
    R_t \odot V^{\mathrm{tgt}}_{t,b}
    +
    (1-R_t) \odot V^{\mathrm{src}}_{t,b}.
\end{aligned}
\label{eq:inject}
\end{equation}
The binary mask $R_t$ is broadcast over feature channels and applied only to image tokens. Features inside $R_t$ remain target-driven, while regions outside $R_t$ are anchored by cached source K/V features to preserve unrelated content. Because $\mathcal{B}_{\mathrm{loc}}$ precedes $\mathcal{B}_{\mathrm{inj}}$, the current edit region is localized before source K/V injection, preventing the injected features from affecting the localization signal.

During the first few denoising steps, we apply global source K/V injection over a short warm-up window $\mathcal{T}_{\mathrm{warm}}$ to anchor the early latent evolution to the source layout and appearance before switching to region-guided K/V injection.

\begin{table*}[t]
\centering
\setlength{\tabcolsep}{5pt}
\small
\begin{tabular}{lcccccccccc}
\toprule
\multirow[c]{2}{*}[-1ex]{Method} &
\multicolumn{5}{c}{\textbf{EditRegion-Bench}} &
\multicolumn{5}{c}{\textbf{PIE-Bench}} \\
\cmidrule(lr){2-6}\cmidrule(lr){7-11}
& AS$\uparrow$ &
$\mathrm{PSNR}\uparrow$ &
$\mathrm{LPIPS}^{\times 10^3}\downarrow$ &
$\text{CLIP}_\text{sim}$$\uparrow$ & ESER$\uparrow$&
AS$\uparrow$ &
$\mathrm{PSNR}\uparrow$ &
$\mathrm{LPIPS}^{\times 10^3}\downarrow$ &
$\text{CLIP}_\text{sim}$$\uparrow$  & ESER$\uparrow$\\
\midrule
PnPInversion  & 6.35 & 24.39 & 80.91 & 27.48 &62.10 & 6.43 & 27.34 & 49.54 & 25.89 & 45.00 \\
MasaCtrl & 6.26 & 22.37 & 87.23 & 26.14 & 27.42& 6.29 & 22.29 & 84.07 & 25.67 & 30.00 \\
RF-Inversion & 6.52 & 20.71 & 181.36 & 27.82 &65.73 & 6.51 & 20.69 & 190.76 & 26.04 & 58.75\\
Stable Flow& 6.52 & 25.85 & 57.06 & 26.44 &27.02 & 6.25 &25.95 & 82.74 & 24.36 & 22.50\\
FlowEdit & 6.54 & 23.63 & 82.38 & 28.18 &70.56 &6.56 & 23.33 & 87.15 & 26.25 & 55.00\\
FYS & 6.59 & 25.77 & 65.53 & 28.22 &73.39 & 6.53 & 24.62 & 116.90 & 25.74 & 60.00\\
RF-Edit & 6.53 & 26.01 & 104.23 & 27.99 &73.39& 6.39 & 25.41 & 127.95 & 25.81 & 62.50\\
\midrule
FLUX.1-Fill$^\dagger$ & 6.47 & 35.43 & 6.18 & 28.30 &81.68 & 6.20 & 36.58 & 11.65 & 25.46 & 66.25\\
KV-Edit$^\dagger$ & 6.52 & 37.47 & 3.97 & 27.41 &65.32 & 6.38 & 36.22 & 8.54 & 24.42 & 50.00\\
\midrule
\textbf{PC-Edit (ours)} & \textbf{6.62} & \textbf{34.37} & \textbf{16.40} & \textbf{28.44} & \textbf{82.44} & \textbf{6.58} & \textbf{29.71} & \textbf{38.92} & \textbf{26.42} & \textbf{67.50}\\
\bottomrule
\end{tabular}
\caption{Editing results on two datasets. $\dagger$ denotes methods given ground-truth masks. Bold indicates the best result among methods without specified edit regions.}
\label{tab:editing}
\end{table*}

\section{Experiments}
\label{sec:exp}

\subsection{Implementation}

We evaluate PC-Edit using FLUX.1-dev~\cite{flux2024} as the backbone on a single NVIDIA A800 GPU. We use 15 inversion steps and 15 denoising steps at a resolution of $1024\times1024$. Following the block and timestep analysis in Fig.~\ref{fig:block_step_analysis}, we set $\mathcal{B}_{\mathrm{loc}}$ to double-stream blocks 13--18 and $\mathcal{B}_{\mathrm{inj}}$ to the last 18 single-stream blocks. For replacement, $M_{\mathrm{src}}$ is estimated from $\mathcal{T}_{\mathrm{probe}}=\{6,7,8\}$, whereas addition sets $M_{\mathrm{src}}=\bm{0}$. The first two denoising steps use global K/V injection ($\mathcal{T}_{\mathrm{warm}}=\{1,2\}$), after which the dynamic mask is updated until $t_s=10$ and then kept fixed.

\begin{table}[t]
\centering
\setlength{\tabcolsep}{2.8pt}
\small
\begin{tabular}{lcccccc}
\toprule
Signal
& AP$\uparrow$
& FBC$\uparrow$
& $\mathrm{bg}_{p99}\downarrow$
& PeakHit$\uparrow$
& mIoU$\uparrow$ \\
\midrule
$\Delta Q$
& 58.2 & 27.6 & 40.8 & 83.5 & 32.0 \\
$\Delta K$
& 60.3 & 26.8 & 39.7 & 86.1 & 33.3 \\
$\Delta V$
& 63.9 & 26.8 & 37.0 & 86.5 & 35.1 \\
$\Delta A$
& 66.6 & 36.5 & 41.9 & 78.2 & 38.6 \\
$\Delta\bm{v}_{\theta}$
& 62.9 & 39.3 & 37.8 & 82.0 & 28.0 \\
Cross-trajectory $\Delta\bm{v}_{\theta}$
& 49.5 & 27.1 & 52.0 & 71.1 & 24.9 \\
\midrule
\textbf{$\bm{\Delta\mathrm{AttnOut}}$ (ours)}
& \textbf{74.4}
& \textbf{44.8}
& \textbf{32.6}
& \textbf{87.2}
& \textbf{40.4} \\
\bottomrule
\end{tabular}
\caption{
Edit-region localization on EditRegion-Bench. All signals except Cross-trajectory $\Delta v_\theta$ use shared-state prompt contrast. Cross-trajectory $\Delta v_\theta$ follows the FYS-style trajectory-divergence setting. Only $\mathrm{bg}_{p99}$ is lower-is-better.
}
\label{tab:localization}
\end{table}

\subsection{Datasets}

We evaluate PC-Edit on EditRegion-Bench and PIE-Bench. EditRegion-Bench contains 484 object-editing cases, comprising 236 additions and 248 replacements, as well as 330 one-object, 87 two-object, and 67 three-object edits. Source images are generated using Seedream~\cite{seedream2025} and GPT-Image-2~\cite{gptimage2} or collected through web image search, and all images are resized to $1024\times1024$. Each case provides source and target prompts together with bounding boxes for all edited objects. The annotations are manually created and independently verified by a second annotator, enabling direct evaluation of both single- and multi-object localization. We additionally use the 240 object-addition and object-replacement cases from PIE-Bench~\cite{ju2024pnpinversion}.

\subsection{Baselines}

We compare PC-Edit against seven mask-free training-free methods. UNet-based baselines include PnPInversion~\citep{ju2024pnpinversion} and MasaCtrl~\citep{cao2023masactrl}. Rectified-flow baselines include RF-Inversion~\citep{rout2025rfinversion}, Stable Flow~\citep{avrahami2025stableflow}, FlowEdit~\citep{kulikov2025flowedit}, FYS~\citep{zhang2025fys}, and RF-Edit~\citep{wang2025taming}. We further report two mask-guided references, FLUX.1-Fill~\citep{flux2024} and KV-Edit~\citep{zhu2025kvedit}, which are given the ground-truth edit boxes as input.

\subsection{Metrics}

Localization is evaluated against the union of the annotated boxes using five metrics. AP evaluates whether edit-region pixels receive higher responses than background pixels~\cite{fan2017structure}. Normalized foreground--background contrast (FBC) quantifies response separation. $\mathrm{bg}_{p99}$ denotes the 99th percentile of responses outside the annotated edit region. PeakHit reports whether the maximum-response pixel falls inside the annotated region. mIoU measures the overlap between the predicted mask and the annotated region \cite{everingham2010pascal}. Editing quality is evaluated using the LAION aesthetic score (AS)~\citep{laionaesthetic} for overall visual quality, CLIP similarity ($\text{CLIP}_\text{sim}$)~\citep{radford2021clip} for semantic alignment with the target prompt, and $\mathrm{PSNR}$ and $\mathrm{LPIPS}$~\citep{zhang2018lpips} for pixel-level and perceptual background fidelity. Background metrics are computed outside the union of the annotated boxes. To directly assess source-object removal, we additionally employ Qwen2.5-VL~\citep{bai2025qwen25} as a frozen automatic evaluator on the replacement cases. We report the Effective Source-Erasure Rate (ESER), which counts a case as successful only when the source concept is no longer recognizable and the annotated edit region undergoes a meaningful change. Detailed prompts, definitions, and auxiliary metrics are provided in the supplementary material.

\subsection{Quantitative Results}
\label{sec:quantitative}

\begin{figure}[t]
    \centering
    \scalebox{1}[0.88]{%
        \includegraphics[width=\columnwidth]{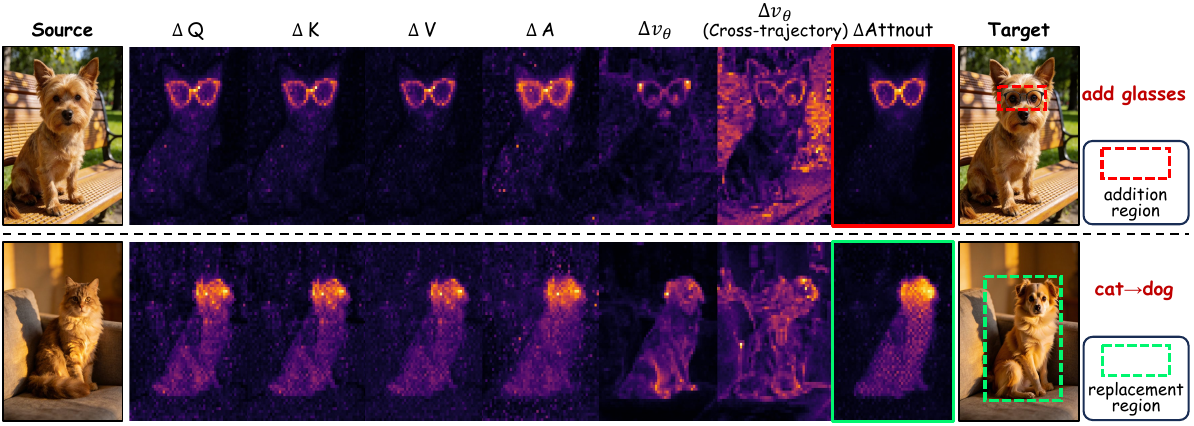}
    }
    \caption{
    Qualitative comparison of edit-region localization signals. $\Delta\mathrm{AttnOut}$ yields the most concentrated responses with less background activation.
    }
    \label{fig:localization_comparison}
\end{figure}

\paragraph{Edit-region localization.}
We compare six candidate readouts for prompt-contrastive localization. For each one, we evaluate the source and target prompts on the same latent at the same timestep and compute the token-wise difference $\Delta X_t=\lVert X_t^{\mathrm{tgt}}-X_t^{\mathrm{src}}\rVert$.
We consider the image-token query $Q$, key $K$, and value $V$ projections, image-to-image attention weights $A$, the image-token attention output $\mathrm{AttnOut}$, and the terminal velocity $\bm{v}_{\theta}$. Except for $\bm{v}_{\theta}$, all signals are extracted from double-stream blocks 13--18. All signals use the same evaluation samples, heatmap normalization, thresholding, and post-processing. We additionally evaluate the FYS-style cross-trajectory velocity difference, which compares the source inversion latent $\bm{x}_t$ with the target denoising latent $\bm{z}_t$ at the same timestep. As shown in Table~\ref{tab:localization}, $\Delta\mathrm{AttnOut}$ achieves the best results across all five metrics. Figure~\ref{fig:localization_comparison} shows the same trend qualitatively, with more concentrated responses and less background activation. The $Q/K/V$ signals are taken from the projections before attention aggregation, whereas $\Delta A$ reflects changes in image-to-image attention distributions without accounting for the value content being aggregated. The terminal velocity is obtained after the prompt-conditioned response has passed through subsequent network computation. Moreover, the lower performance of the FYS-style cross-trajectory signal suggests additional interference from comparing different latent states. Overall, these results identify $\Delta\mathrm{AttnOut}$ as the most reliable readout among the compared signals for edit-region localization.

\paragraph{Editing quality.} Table~\ref{tab:editing} shows that PC-Edit achieves the best overall balance between editability and content preservation among methods without externally specified edit regions. It provides the highest background fidelity across both benchmarks, while also attaining the best aesthetic quality and target-prompt alignment. Compared with the GT-box-guided references, PC-Edit substantially narrows the preservation gap without access to ground-truth boxes, while producing more visually appealing and semantically faithful edits. A complementary Qwen2.5-VL-based evaluation on replacement cases further shows that PC-Edit achieves the highest ESER score among all methods, indicating effective source removal alongside strong background preservation. Full evaluator details and auxiliary metrics are provided in the supplementary material.

\begin{table}[t]
\centering
\setlength{\tabcolsep}{3.5pt}
\small
\begin{tabular}{lcccc}
\toprule
Settings
& AS$\uparrow$
& $\mathrm{PSNR}\uparrow$
& $\mathrm{LPIPS}^{\times 10^3}\downarrow$
& CLIP$\uparrow$
 \\
\midrule
w/o $M_{\mathrm{src}}$
& 6.50 & \textbf{34.11} & \textbf{18.87} & 28.00 \\

Static target mask
& 6.54 & 30.55 & 31.12 & 28.17 \\

Staged control
& 6.55 & 26.61 & 50.78 & 28.13 \\

\textbf{Full PC-Edit (ours)}
& \textbf{6.57} & \underline{33.25} & \underline{19.50} & \textbf{28.21} \\
\bottomrule
\end{tabular}
\caption{
Component ablation on the 248 replacement cases of EditRegion-Bench.
}
\label{tab:ablation}
\end{table}

\subsection{Qualitative Results}
\label{sec:exp-qualitative}

\begin{figure}[t]
    \centering
    \scalebox{1}[0.89]{%
        \includegraphics[width=\columnwidth]{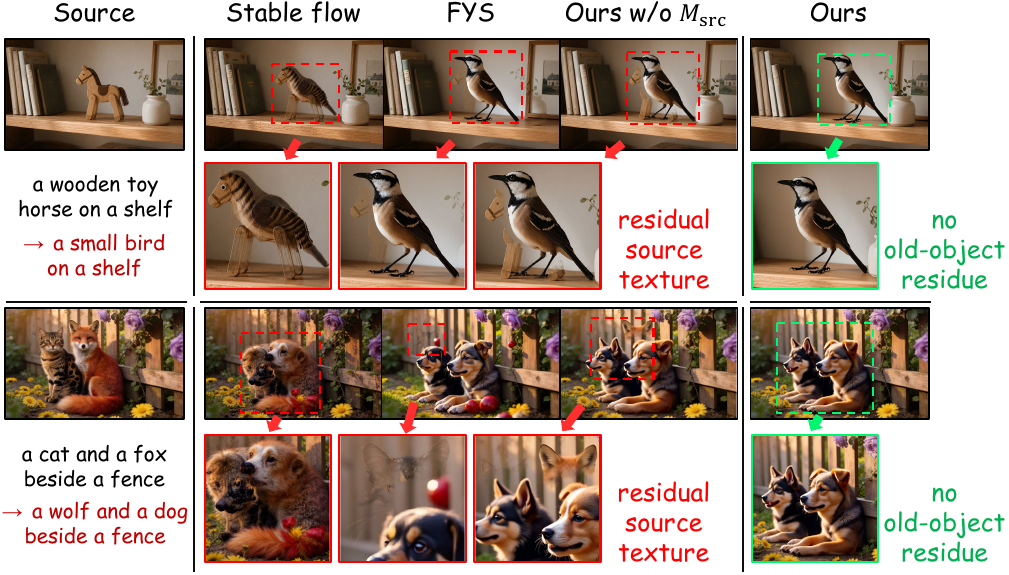}
    }
    \caption{
    Effect of source-erasure modeling. Explicit $M_{\mathrm{src}}$ suppresses residual source textures observed in competing methods and the variant without $M_{\mathrm{src}}$.
    }
    \label{fig:source_erasure}
\end{figure}

Figure~\ref{fig:qualitative_comparison} presents qualitative comparisons on single- and multi-object addition and replacement. PC-Edit successfully performs both small localized edits and large cross-category replacements while preserving unrelated regions. Compared with mask-free baselines, it exhibits fewer missed edits, source remnants, and unintended background changes.
Figure~\ref{fig:source_erasure} further shows that explicitly modeling $M_{\mathrm{src}}$ suppresses residual source content. Although GT-box-guided methods preserve the background well, their predefined regions may restrict target formation when the source and target have different silhouettes. By jointly modeling the source-erasure region and tracking target emergence during denoising, PC-Edit allows the target to extend beyond the source footprint while preserving unrelated content.

\begin{figure}[t]
    \centering
    \scalebox{1}[0.975]{\includegraphics[
        width=\columnwidth,
        pagebox=cropbox
    ]{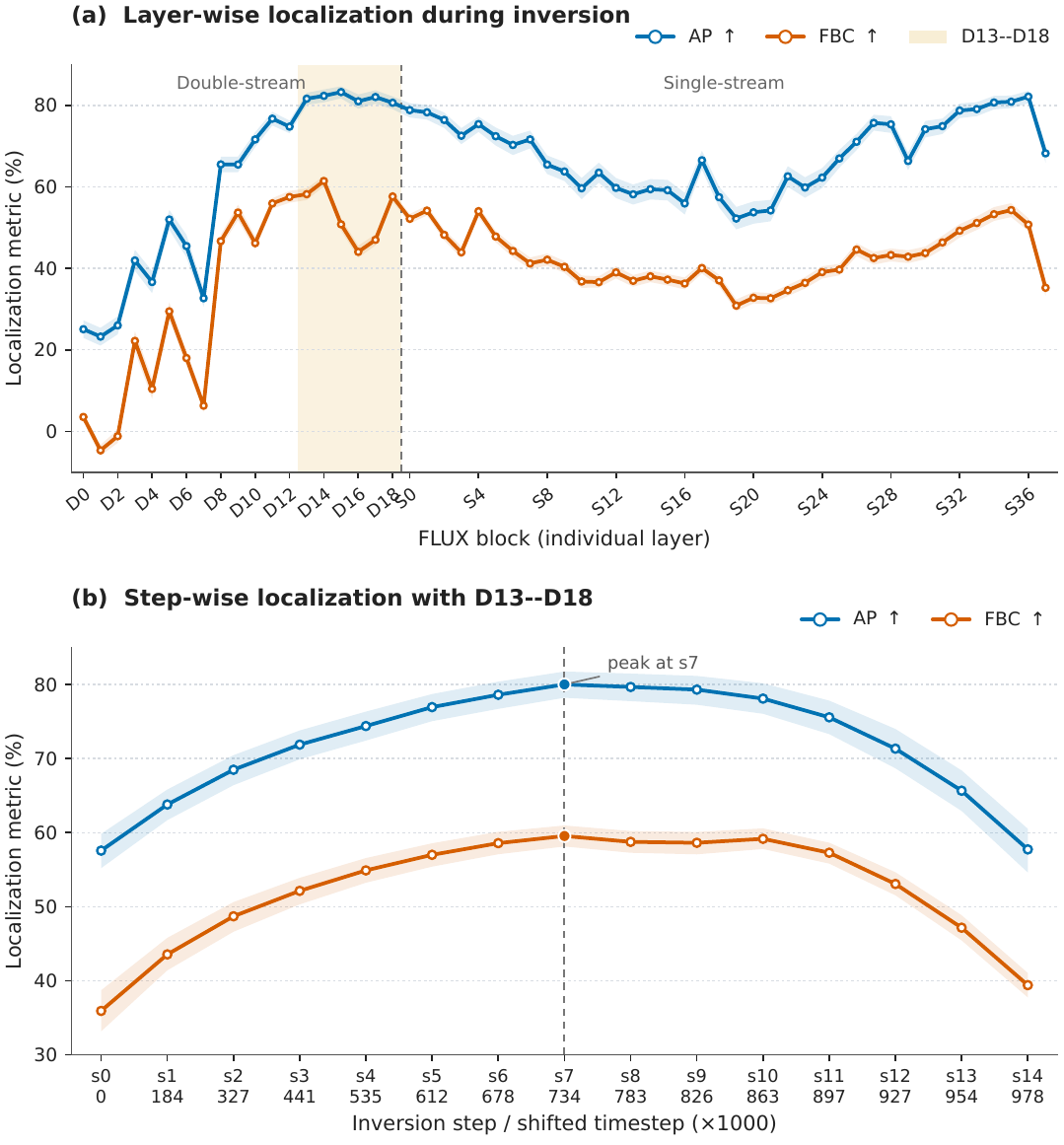}}
    \caption{
    Localization across FLUX blocks and inversion steps.
    (a) Per-block AP and FBC averaged over steps.
    (b) Per-step results using double-stream blocks 13--18.
    }
    \label{fig:block_step_analysis}
\end{figure}

\subsection{Ablation Study}
\label{sec:exp-ablation}

We first examine the architectural and temporal design choices for region discovery, followed by component-wise ablations of PC-Edit.

\paragraph{Block and inversion-step selection.}
Figure~\ref{fig:block_step_analysis} evaluates $\Delta\mathrm{AttnOut}$ across FLUX blocks and inversion steps. Averaged over inversion steps, double-stream blocks 13--18 yield the strongest and most stable localization. With these blocks fixed, both AP and FBC peak in the middle of inversion. Therefore, we set
$\mathcal{B}_{\mathrm{loc}}=\{13,\ldots,18\}$ and
$\mathcal{T}_{\mathrm{probe}}=\{6,7,8\}$ for constructing
$M_{\mathrm{src}}$.

\paragraph{Component ablation.}
We evaluate the main components of PC-Edit on the 248 replacement cases of EditRegion-Bench, where source removal is explicitly required. We consider three variants. \emph{w/o $M_{\mathrm{src}}$} removes the inversion-time source-erasure footprint and defines the edit region solely from the denoising-time target-emergence mask. \emph{Static target mask} estimates the target-emergence region once after the global-injection warm-up and reuses it throughout the remaining denoising steps, thereby removing online edit region refinement. \emph{Staged control} follows an FYS-style schedule, allowing the latent to denoise freely under the target prompt while accumulating the edit mask, and applying source K/V injection outside the resulting mask only during the final denoising steps. As shown in Table~\ref{tab:ablation}, the variants exhibit different trade-offs between source removal and background preservation. Although removing $M_{\mathrm{src}}$ slightly improves the background-fidelity metrics, it leaves more source remnants because the target-emergence region may not fully cover the source footprint, particularly when the source and target silhouettes differ substantially. The Qwen-based evaluation confirms this limitation: adding $M_{\mathrm{src}}$ increases ESER from 72.18\% to 82.44\%, while Figure~\ref{fig:source_erasure} qualitatively shows residual source content in the variant without $M_{\mathrm{src}}$. Staged control degrades background fidelity, as background drift accumulated during early unconstrained denoising cannot be fully corrected by source K/V injection in the final steps. The static target mask is estimated only once after warm-up, while the target response is still relatively diffuse, and remains fixed thereafter. It therefore cannot adapt as the target structure emerges, often resulting in an overly broad edit region and weakened background preservation. Delaying mask collection could yield a more accurate region estimate, but would leave the preceding denoising steps unconstrained and inherit the same limitation as staged control. In contrast, PC-Edit refines the target-emergence region online after warm-up and immediately applies region-guided control within the same sampling step, maintaining edit coverage while limiting background drift.

\section{Conclusion}
\label{sec:conclusion}

We presented PC-Edit, a training-free framework for prompt-contrastive region discovery in MM-DiT editing. By holding the latent state and timestep fixed while varying only the prompt, PC-Edit captures prompt-induced semantic differences from image-token attention outputs. Applying the same contrast during inversion and denoising produces source-erasure and target-emergence regions, whose union defines the edit region. Within each sampling step, PC-Edit immediately injects cached source K/V features outside this region, preserving unrelated content before the latent update. We introduced EditRegion-Bench, which provides human-verified edit-region annotations for single- and multi-object addition and replacement. Experiments on EditRegion-Bench and PIE-Bench demonstrate that PC-Edit achieves more accurate edit-region localization and the best balance of editing quality and background preservation among methods that do not require manually specified edit regions.

\bibliography{aaai2027}

@inproceedings{avrahami2025stableflow,
  title={Stable flow: Vital layers for training-free image editing},
  author={Avrahami, Omri and Patashnik, Or and Fried, Ohad and Nemchinov, Egor and Aberman, Kfir and Lischinski, Dani and Cohen-Or, Daniel},
  booktitle={Proceedings of the Computer Vision and Pattern Recognition Conference},
  pages={7877--7888},
  year={2025}
}

@inproceedings{cao2023masactrl,
  title={Masactrl: Tuning-free mutual self-attention control for consistent image synthesis and editing},
  author={Cao, Mingdeng and Wang, Xintao and Qi, Zhongang and Shan, Ying and Qie, Xiaohu and Zheng, Yinqiang},
  booktitle={Proceedings of the IEEE/CVF international conference on computer vision},
  pages={22560--22570},
  year={2023}
}

@inproceedings{couairon2023diffedit,
  title={DiffEdit: Diffusion-based Semantic Image Editing with Mask Guidance},
  author={Couairon, Guillaume and Verbeek, Jakob and Schwenk, Holger and Cord, Matthieu},
  booktitle={ICLR 2023 (Eleventh International Conference on Learning Representations)},
  year={2023}
}

@inproceedings{esser2024sd3,
  title={Scaling rectified flow transformers for high-resolution image synthesis},
  author={Esser, Patrick and Kulal, Sumith and Blattmann, Andreas and Entezari, Rahim and M{\"u}ller, Jonas and Saini, Harry and Levi, Yam and Lorenz, Dominik and Sauer, Axel and Boesel, Frederic and others},
  booktitle={Forty-first international conference on machine learning},
  year={2024}
}

@article{hertz2023prompt,
  title={Prompt-to-Prompt Image Editing with Cross Attention Control},
  author={Hertz, Amir and Mokady, Ron and Tenenbaum, Jay and Aberman, Kfir and Pritch, Yael and Cohen-Or, Daniel},
  journal={arXiv e-prints},
  pages={arXiv--2208},
  year={2022}
}

@article{everingham2010pascal,
  title={The pascal visual object classes (voc) challenge},
  author={Everingham, Mark and Van Gool, Luc and Williams, Christopher KI and Winn, John and Zisserman, Andrew},
  journal={International journal of computer vision},
  volume={88},
  number={2},
  pages={303--338},
  year={2010},
  publisher={Springer}
}

@inproceedings{ju2024pnpinversion,
  title={Pnp inversion: Boosting diffusion-based editing with 3 lines of code},
  author={Ju, Xuan and Zeng, Ailing and Bian, Yuxuan and Liu, Shaoteng and Xu, Qiang},
  booktitle={International Conference on Learning Representations},
  volume={2024},
  pages={23395--23422},
  year={2024}
}

@inproceedings{kulikov2025flowedit,
  title={Flowedit: Inversion-free text-based editing using pre-trained flow models},
  author={Kulikov, Vladimir and Kleiner, Matan and Huberman-Spiegelglas, Inbar and Michaeli, Tomer},
  booktitle={Proceedings of the IEEE/CVF International Conference on Computer Vision},
  pages={19721--19730},
  year={2025}
}

@inproceedings{lipman2023flow,
  title={Flow Matching for Generative Modeling},
  author={Lipman, Yaron and Chen, Ricky TQ and Ben-Hamu, Heli and Nickel, Maximilian and Le, Matt},
  booktitle={11th International Conference on Learning Representations, ICLR 2023},
  year={2023}
}

@inproceedings{liu2023rectified,
  title={Flow Straight and Fast: Learning to Generate and Transfer Data with Rectified Flow},
  author={Liu, Xingchao and Gong, Chengyue and Liu, Qiang},
  booktitle={The Eleventh International Conference on Learning Representations (ICLR)},
  year={2023}
}

@inproceedings{mokady2023nulltext,
  title={Null-text inversion for editing real images using guided diffusion models},
  author={Mokady, Ron and Hertz, Amir and Aberman, Kfir and Pritch, Yael and Cohen-Or, Daniel},
  booktitle={Proceedings of the IEEE/CVF conference on computer vision and pattern recognition},
  pages={6038--6047},
  year={2023}
}

@inproceedings{peebles2023dit,
  title={Scalable diffusion models with transformers},
  author={Peebles, William and Xie, Saining},
  booktitle={Proceedings of the IEEE/CVF international conference on computer vision},
  pages={4195--4205},
  year={2023}
}

@inproceedings{radford2021clip,
  title={Learning transferable visual models from natural language supervision},
  author={Radford, Alec and Kim, Jong Wook and Hallacy, Chris and Ramesh, Aditya and Goh, Gabriel and Agarwal, Sandhini and Sastry, Girish and Askell, Amanda and Mishkin, Pamela and Clark, Jack and others},
  booktitle={International conference on machine learning},
  pages={8748--8763},
  year={2021},
  organization={PmLR}
}

@inproceedings{feng2025dit4edit,
  title={Dit4edit: Diffusion transformer for image editing},
  author={Feng, Kunyu and Ma, Yue and Wang, Bingyuan and Qi, Chenyang and Chen, Haozhe and Chen, Qifeng and Wang, Zeyu},
  booktitle={Proceedings of the AAAI Conference on Artificial Intelligence},
  volume={39},
  number={3},
  pages={2969--2977},
  year={2025}
}

@inproceedings{wang2025taming,
  title={Taming Rectified Flow for Inversion and Editing},
  author={Wang, Jiangshan and Pu, Junfu and Qi, Zhongang and Guo, Jiayi and Ma, Yue and Huang, Nisha and Chen, Yuxin and Li, Xiu and Shan, Ying},
  booktitle={International Conference on Machine Learning},
  pages={64044--64058},
  year={2025},
  organization={PMLR}
}

@inproceedings{rombach2022ldm,
  title={High-resolution image synthesis with latent diffusion models},
  author={Rombach, Robin and Blattmann, Andreas and Lorenz, Dominik and Esser, Patrick and Ommer, Bj{\"o}rn},
  booktitle={Proceedings of the IEEE/CVF conference on computer vision and pattern recognition},
  pages={10684--10695},
  year={2022}
}

@inproceedings{
    rout2025rfinversion,
    title={Semantic Image Inversion and Editing using Rectified Stochastic Differential Equations},
    author={Litu Rout and Yujia Chen and Nataniel Ruiz and Constantine Caramanis and Sanjay Shakkottai and Wen-Sheng Chu},
    booktitle={The Thirteenth International Conference on Learning Representations},
    year={2025},
    url={https://openreview.net/forum?id=Hu0FSOSEyS}
}

@article{seedream2025,
  title={Seedream 3.0 technical report},
  author={Gao, Yu and Gong, Lixue and Guo, Qiushan and Hou, Xiaoxia and Lai, Zhichao and Li, Fanshi and Li, Liang and Lian, Xiaochen and Liao, Chao and Liu, Liyang and others},
  journal={arXiv preprint arXiv:2504.11346},
  year={2025}
}

@inproceedings{tumanyan2023pnp,
  title={Plug-and-play diffusion features for text-driven image-to-image translation},
  author={Tumanyan, Narek and Geyer, Michal and Bagon, Shai and Dekel, Tali},
  booktitle={Proceedings of the IEEE/CVF conference on computer vision and pattern recognition},
  pages={1921--1930},
  year={2023}
}

@misc{gptimage2,
  author       = {{OpenAI}},
  title        = {{GPT Image 2}},
  year         = {2026},
  howpublished = {\url{https://developers.openai.com/api/docs/models/gpt-image-2}}
}

@inproceedings{fan2017structure,
  title={Structure-measure: A new way to evaluate foreground maps},
  author={Fan, Deng-Ping and Cheng, Ming-Ming and Liu, Yun and Li, Tao and Borji, Ali},
  booktitle={Proceedings of the IEEE international conference on computer vision},
  pages={4548--4557},
  year={2017}
}

@article{otsu1975threshold,
  title={A threshold selection method from gray-level histograms},
  author={Otsu, Nobuyuki},
  journal={Automatica},
  volume={11},
  pages={285--296},
  year={1975}
}

@article{vaswani2017attention,
  title={Attention is all you need},
  author={Vaswani, Ashish and Shazeer, Noam and Parmar, Niki and Uszkoreit, Jakob and Jones, Llion and Gomez, Aidan N and Kaiser, {\L}ukasz and Polosukhin, Illia},
  journal={Advances in neural information processing systems},
  volume={30},
  year={2017}
}

@inproceedings{zhang2018lpips,
  title={The unreasonable effectiveness of deep features as a perceptual metric},
  author={Zhang, Richard and Isola, Phillip and Efros, Alexei A and Shechtman, Eli and Wang, Oliver},
  booktitle={Proceedings of the IEEE conference on computer vision and pattern recognition},
  pages={586--595},
  year={2018}
}

@inproceedings{zhu2025kvedit,
  title={Kv-edit: Training-free image editing for precise background preservation},
  author={Zhu, Tianrui and Zhang, Shiyi and Shao, Jiawei and Tang, Yansong},
  booktitle={Proceedings of the IEEE/CVF International Conference on Computer Vision},
  pages={16607--16617},
  year={2025}
}

@misc{flux2024,
  author = {{Black Forest Labs}},
  title = {{FLUX.}},
  year = {2024},
  howpublished = {\url{https://github.com/black-forest-labs/flux}}
}

@article{laionaesthetic,
  title={Laion-5b: An open large-scale dataset for training next generation image-text models},
  author={Schuhmann, Christoph and Beaumont, Romain and Vencu, Richard and Gordon, Cade and Wightman, Ross and Cherti, Mehdi and Coombes, Theo and Katta, Aarush and Mullis, Clayton and Wortsman, Mitchell and others},
  journal={Advances in neural information processing systems},
  volume={35},
  pages={25278--25294},
  year={2022}
}

@article{bai2025qwen25,
  title={Qwen2.5-VL Technical Report},
  author={Shuai Bai and Keqin Chen and Xuejing Liu and Jialin Wang and Wenbin Ge and Sibo Song and Kai Dang and Peng Wang and Shijie Wang and Jun Tang and Humen Zhong and Yuanzhi Zhu and Mingkun Yang and Zhaohai Li and Jianqiang Wan and Pengfei Wang and Wei Ding and Zheren Fu and Yiheng Xu and Jiabo Ye and Xi Zhang and Tianbao Xie and Zesen Cheng and Hang Zhang and Zhibo Yang and Haiyang Xu and Junyang Lin},
  journal={arXiv preprint arXiv:2511.21631},
  year={2025}
}

@inproceedings{
zhang2025fys,
title={Follow-Your-Shape: Shape-Aware Image Editing via Trajectory-Guided Region Control},
author={Zeqian Long and Mingzhe Zheng and Kunyu Feng and Xinhua Zhang and Hongyu Liu and Harry Yang and Linfeng Zhang and Qifeng Chen and Yue Ma},
booktitle={The Fourteenth International Conference on Learning Representations},
year={2026},
url={https://openreview.net/forum?id=uGaR7L3Z1E}
}

\clearpage

\twocolumn[
\begin{@twocolumnfalse}
\begin{center}
    {\LARGE\bfseries Supplementary Material\par}
\end{center}

\vspace{1em}
\end{@twocolumnfalse}
]

\vspace{1em}

This supplement presents the complete PC-Edit inference procedure and fixed configuration, further details of EditRegion-Bench, metric definitions, Qwen2.5-VL-based source-erasure evaluation, additional qualitative results, and the complete evaluator prompts. Unless otherwise stated, notation follows the main paper.

\section{Inference Procedure and Configuration}

Algorithm~\ref{alg:pcedit-inversion} and~\ref{alg:pcedit-denoising} summarize the inference procedure of PC-Edit. During inversion, source-conditioned forward passes advance the source trajectory and cache the corresponding K/V features. For object replacement, additional target-conditioned passes are performed at the inversion latents indexed by $\mathcal{T}_{\mathrm{probe}}$ to estimate the source-erasure mask $M_{\mathrm{src}}$. Addition skips this stage and sets $M_{\mathrm{src}}=\bm{0}$. During denoising, prompt-contrastive localization is performed in $\mathcal{B}_{\mathrm{loc}}$ before source K/V injection in $\mathcal{B}_{\mathrm{inj}}$, allowing the region estimated at the current step to guide the same latent update. Global source K/V injection is applied during the warm-up steps $\mathcal{T}_{\mathrm{warm}}$. The edit region is subsequently updated until the freeze step $t_s$ and kept fixed. Table~\ref{tab:implementation} lists the fixed configuration used in all main-paper experiments.

\section{EditRegion-Bench Details}

EditRegion-Bench contains 484 evaluation cases: 236 additions and 248 replacements, covering one-, two-, and three-object edits as reported in the main paper. Each case includes a $1024\times1024$ source image, a source prompt, a target prompt, and one human-annotated bounding box for each requested edit target. Localization evaluation uses the union of these boxes. The annotations were drawn by one annotator and independently verified by a second. They mark the image regions expected to change under the editing instruction and are used only for evaluation, without being provided to PC-Edit or any other mask-free method during inference. Since valid edits may occur at different reasonable locations and cover different areas, our annotations indicate where the edit should happen rather than define an exact pixel-level mask. Figures~\ref{fig:addition-data} and~\ref{fig:replacement-data} show representative examples from EditRegion-Bench, where the red overlays visualize the corresponding human annotations. EditRegion-Bench is publicly available at \url{https://zenodo.org/records/21502501}.

\begin{table}[H]
\centering
\tiny
\setlength{\tabcolsep}{4.0pt}
\begin{tabular}{@{}p{0.50\columnwidth}p{0.45\columnwidth}@{}}
\toprule
\textbf{Component} & \textbf{Setting} \\
\midrule
Backbone & FLUX.1-dev \\
Image resolution & $1024 \times 1024$ \\
Inversion / denoising steps & $15 / 15$ \\
Guidance scale & $3.0$ \\
Localization blocks $\mathcal B_{\mathrm{loc}}$
    & Double-stream blocks 13--18 \\
Injection blocks $\mathcal B_{\mathrm{inj}}$
    & Last 18 single-stream blocks \\
Probing steps $\mathcal T_{\mathrm{probe}}$
    & $\{6,7,8\}$ \\
Source-erasure mask for addition
    & $M_{\mathrm{src}}=\bm{0}$ \\
Mask binarization & Otsu thresholding \\
Warm-up steps $\mathcal T_{\mathrm{warm}}$
    & $\{1,2\}$ \\
Region-update steps & $3$--$10$ \\
Region-freeze step $t_s$ & $10$ \\
Hardware & One NVIDIA A800 GPU \\
\bottomrule
\end{tabular}
\caption{Fixed inference configuration of PC-Edit.}
\label{tab:implementation}
\end{table}

\begin{algorithm}[!t]
\caption{PC-Edit source inversion}
\fontsize{8pt}{7pt}\selectfont
\label{alg:pcedit-inversion}
\begin{algorithmic}[1]
\STATE \textbf{Input:} Source image $I_{\src}$, conditions
$(\bm c_{\src},\bm c_{\tgt})$, edit type $e$,
$\mathcal B_{\mathrm{loc}}$, $\mathcal B_{\mathrm{inj}}$,
and $\mathcal T_{\mathrm{probe}}$
\STATE \textbf{Output:} Inverted latent $\bm x_N$, cached source
features $\{\mathcal C_t\}_{t=1}^{N}$, and source-erasure mask
$M_{\src}$

\STATE $\bm x_0\leftarrow\operatorname{Encode}(I_{\src})$
\STATE $\mathcal H_{\src}\leftarrow\varnothing$

\FOR{$t=1$ \TO $N$}
    \STATE Run the source-conditioned forward pass to obtain
    $\bm v_t^{\src}$ and image-token attention outputs
    $\mathrm{AttnOut}_t^{\src}$ from $\mathcal B_{\mathrm{loc}}$
    \STATE Cache the source K/V features
    $\mathcal C_t=\{K_{t,b}^{\src},V_{t,b}^{\src}\}_{b\in\mathcal B_{\mathrm{inj}}}$

    \IF{$e=\textsc{replacement}$ \AND
    $t\in\mathcal T_{\mathrm{probe}}$}
        \STATE Obtain $\mathrm{AttnOut}_t^{\tgt}$ from
        $\mathcal B_{\mathrm{loc}}$ at
        $(\bm x_{t-1},\tau_t,\bm c_{\tgt})$
        \STATE $H_t\leftarrow
        |\mathcal B_{\mathrm{loc}}|^{-1}
        \sum_{b\in\mathcal B_{\mathrm{loc}}}
        \|\mathrm{AttnOut}_{t,b}^{\tgt}
        -\mathrm{AttnOut}_{t,b}^{\src}\|$
        \STATE Append $H_t$ to $\mathcal H_{\src}$
    \ENDIF

    \STATE $\bm x_t\leftarrow
    \operatorname{InvertStep}
    (\bm x_{t-1},\bm v_t^{\src})$
\ENDFOR

\IF{$e=\textsc{replacement}$}
    \STATE $M_{\src}\leftarrow
    \operatorname{Otsu}
    (\operatorname{Mean}(\mathcal H_{\src}))$
\ELSE
    \STATE $M_{\src}\leftarrow\bm 0$
\ENDIF

\STATE \textbf{return}
$\bm x_N$, $\{\mathcal C_t\}_{t=1}^{N}$, and $M_{\src}$
\end{algorithmic}
\end{algorithm}

\begin{algorithm}[H]
\caption{PC-Edit target denoising}
\fontsize{8pt}{7pt}\selectfont
\label{alg:pcedit-denoising}
\begin{algorithmic}[1]
\STATE \textbf{Input:} Inverted latent $\bm x_N$, conditions
$(\bm c_{\src},\bm c_{\tgt})$, cached source features
$\{\mathcal C_t\}_{t=1}^{N}$, source-erasure mask $M_{\src}$,
$\mathcal B_{\mathrm{loc}}$, $\mathcal B_{\mathrm{inj}}$,
$\mathcal T_{\mathrm{warm}}$, and $t_s$
\STATE \textbf{Output:} Edited image $I_{\mathrm{edit}}$

\STATE $\bm z_N\leftarrow\bm x_N$
\STATE $R\leftarrow M_{\src}$

\FOR{$k=1$ \TO $N$}
    \STATE $t\leftarrow N-k+1$

    \IF{$k\in\mathcal T_{\mathrm{warm}}$}
        \STATE Run the target-conditioned forward pass using cached
        source K/V for all image tokens to obtain $\bm v_t$
    \ELSE
        \STATE Run the target-conditioned forward pass through
        $\mathcal B_{\mathrm{loc}}$ to obtain
        $\mathrm{AttnOut}_t^{\tgt}$

        \IF{$k\leq t_s$}
            \STATE Obtain $\mathrm{AttnOut}_t^{\src}$ from
            $\mathcal B_{\mathrm{loc}}$ at
            $(\bm z_t,\tau_t,\bm c_{\src})$
            \STATE $H_t\leftarrow
            |\mathcal B_{\mathrm{loc}}|^{-1}
            \sum_{b\in\mathcal B_{\mathrm{loc}}}
            \|\mathrm{AttnOut}_{t,b}^{\tgt}
            -\mathrm{AttnOut}_{t,b}^{\src}\|$
            \STATE $R\leftarrow
            M_{\src}\cup\operatorname{Otsu}(H_t)$
        \ENDIF

        \STATE Continue the target-conditioned forward pass through
        $\mathcal B_{\mathrm{inj}}$ using target K/V inside $R$
        and cached source K/V outside $R$ to obtain $\bm v_t$
    \ENDIF

    \STATE $\bm z_{t-1}\leftarrow
    \operatorname{DenoiseStep}
    (\bm z_t,\bm v_t)$
\ENDFOR

\STATE \textbf{return}
$I_{\mathrm{edit}}\leftarrow
\operatorname{Decode}(\bm z_0)$
\end{algorithmic}
\end{algorithm}

\begin{figure*}[!t]
\centering
\includegraphics[width=\textwidth]{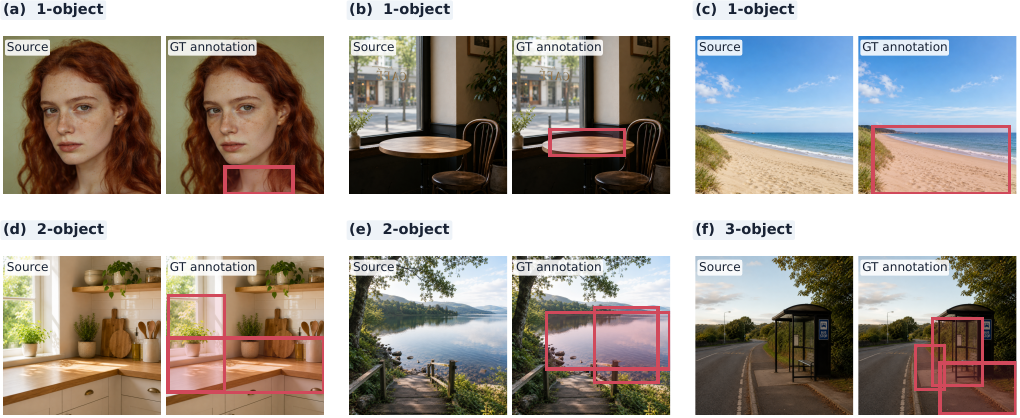}
\setlength{\tabcolsep}{3.5pt}
\begin{tabular}{cp{3.08in}p{3.08in}}
\toprule
Panel & Source prompt & Target prompt \\
\midrule
(a) & a portrait of a freckled woman with wavy auburn hair, against a light olive backdrop & a portrait of a freckled woman with wavy auburn hair, against a light olive backdrop, wearing an amber bead necklace \\
(b) & an empty café table near a window & an empty café table near a window with a cup of coffee on it \\
(c) & a sandy beach near the ocean & a sandy beach near the ocean with a colorful beach ball on it \\
(d) & a kitchen counter near a sunny window & a kitchen counter near a sunny window with a bowl of oranges and a glass of milk on it \\
(e) & a quiet lake shore with wooden steps & a quiet lake shore with wooden steps, a small boat, and a duck on the water \\
(f) & a calm riverside walkway with benches & a calm riverside walkway with benches, a parked bicycle, a person sitting on a bench, and a small dog nearby \\
\bottomrule
\end{tabular}
\caption{Representative EditRegion-Bench addition cases and their prompts.  Each panel pairs the source image with the human edit-region annotation.  Panels (a)--(c) are single-object additions, (d)--(e) are two-object additions, and (f) is a three-object addition.}
\label{fig:addition-data}
\end{figure*}

\begin{figure*}[!t]
\centering
\includegraphics[width=\textwidth]{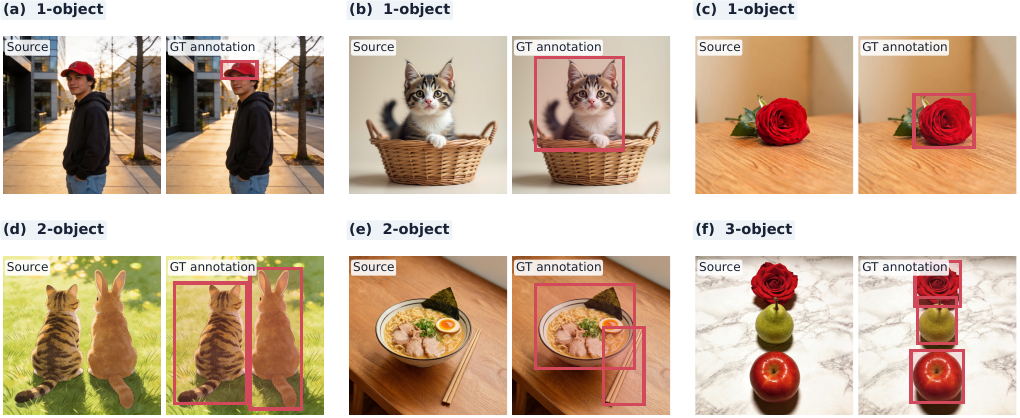}
\setlength{\tabcolsep}{3.5pt}
\begin{tabular}{cp{3.08in}p{3.08in}}
\toprule
Panel & Source prompt & Target prompt \\
\midrule
(a) & a young man wearing a red baseball cap, standing on a city sidewalk, lit by soft daylight & a young man wearing a blue baseball cap, standing on a city sidewalk, lit by soft daylight \\
(b) & a photo of a cat in a basket & a photo of a dog in a basket \\
(c) & a red rose on a wooden table & a pink tulip on a wooden table \\
(d) & a tabby cat and a brown rabbit sitting in a row on green grass & a brown hare and a small dog sitting in a row on green grass \\
(e) & a wooden table set with a bowl of ramen and chopsticks, photographed from a slightly higher angle & a wooden table set with a bowl of salad and a fork, photographed from a slightly higher angle \\
(f) & a red apple, a green pear, and a red rose arranged in a row on a marble counter & an orange, a ripe mango, and a pink tulip arranged in a row on a marble counter \\
\bottomrule
\end{tabular}
\caption{Representative EditRegion-Bench replacement cases and their prompts.  Panels (a)--(c) are single-object replacements, (d)--(e) are two-object replacements, and (f) is a three-object replacement.  Each annotated view shows the union's constituent human boxes before unioning.}
\label{fig:replacement-data}
\end{figure*}

\section{Metric Definition}
\label{sec:metrics}

\subsection{Localization Metrics}
Let $H$ be a heatmap, where $H(p)$ denotes the response at pixel $p$,
and let $G$ denote the set of pixels covered by the union of the
human-annotated boxes. Each heatmap is independently normalized to
$[0,1]$:
\begin{equation}
\widehat H(p)=
\begin{cases}
\dfrac{H(p)-H_{\min}}{H_{\max}-H_{\min}}, & H_{\max}>H_{\min},\\[0.35em]
0, & H_{\max}=H_{\min},
\end{cases}
\end{equation}
where $H_{\min}$ and $H_{\max}$ are the minimum and maximum responses
over all pixels, respectively.

\paragraph{Average precision (AP).}
Pixels in $G$ are treated as positives and all remaining pixels as
negatives. We sort all pixels in descending order of their normalized
heatmap responses and compute the standard non-interpolated average
precision:
\begin{equation}
\mathrm{AP}=\sum_k (R_k-R_{k-1})P_k,
\end{equation}
where $P_k$ and $R_k$ are the precision and recall at the $k$th point
on the precision--recall curve. AP evaluates whether annotated-region pixels receive higher responses.

\paragraph{Foreground--background contrast (FBC).}
Let $N$ denote the total number of pixels. We first compute the mean responses inside and outside the annotated region:
\begin{equation}
\mu_{\mathrm{fg}}
=
\frac{1}{|G|}
\sum_{p\in G}\widehat H(p),
\qquad
\mu_{\mathrm{bg}}
=
\frac{1}{N-|G|}
\sum_{p\notin G}\widehat H(p).
\end{equation}
FBC is then defined as
\begin{equation}
\mathrm{FBC}
=
\frac{\mu_{\mathrm{fg}}-\mu_{\mathrm{bg}}}
{\mu_{\mathrm{fg}}+\mu_{\mathrm{bg}}+\eps}.
\label{eq:fbc}
\end{equation}
We use $\eps=10^{-8}$. FBC lies in $[-1,1]$, with higher values indicating stronger separation between annotated and background regions.

\paragraph{Background 99th percentile ($\mathrm{bg}_{p99}$) and PeakHit.}
The high-background response is defined as
\begin{equation}
\mathrm{bg}_{p99}
=
Q_{0.99}
\bigl(
\{\widehat H(p)|p\notin G\}
\bigr),
\end{equation}
where $Q_{0.99}$ denotes the 99th percentile and lower values are better. PeakHit is defined as
\begin{equation}
\mathrm{PeakHit}
=
\arg\max_p \widehat H(p)\in G,
\end{equation}
which indicates whether the strongest heatmap response falls inside an annotated edit region.

\paragraph{Mask IoU (mIoU).}
We apply Otsu's threshold to each heatmap to obtain a binary mask $M$.
The per-case IoU is
\begin{equation}
\mathrm{IoU}
=
\frac{|M\cap G|}
{|M\cup G|+\eps},
\end{equation}
where $\eps=10^{-8}$. mIoU is the mean IoU over all cases.

\subsection{Editing Metrics}

\paragraph{Aesthetic score (AS).}
The aesthetic quality of the edited images is evaluated using the LAION aesthetic score.

\paragraph{CLIP score ($\text{CLIP}_\text{sim}$).}
We use TorchMetrics $\text{CLIP}_\text{sim}$ with \texttt{openai/clip-vit-large-patch14} to measure the image--text similarity between $I_{\mathrm{edit}}$ and the target prompt $\bm c_{\tgt}$. The score is reported on its native 0--100 scale, with higher values indicating stronger semantic alignment.

\paragraph{Background fidelity.}
We evaluate background preservation using $\mathrm{PSNR}$ and $\mathrm{LPIPS}$. Let $B$ denote the background binary mask.  Background PSNR is computed from the masked edited and source images:
\begin{equation}
\mathrm{PSNR}
=
-10\log_{10}
\left(
\operatorname{MSE}
\left(
B\odot I_{\mathrm{edit}},
B\odot I_{\src}
\right)
\right),
\end{equation}
where the image range is $[0,1]$. The MSE is computed over the complete masked images, so pixels inside the annotated edit region are set to zero in both images.

For $\mathrm{LPIPS}$, the same masked images are mapped from $[0,1]$ to $[-1,1]$ and evaluated using the SqueezeNet-based LPIPS network. The tables report $\mathrm{LPIPS}\times10^{3}$ where indicated.

\paragraph{Qwen-based source-erasure evaluation.}
To directly evaluate whether the source concept is effectively removed, we additionally employ Qwen2.5-VL-7B-Instruct as a frozen automatic evaluator on replacement cases.

For each edited result, the evaluator first determines whether the source concept or any recognizable source-specific cue remains in the edited image, including characteristic parts, colors, textures, materials, or silhouettes. It then compares the source and edited images to determine whether the annotated edit region undergoes a meaningful visual or semantic change.

For the $i$-th edited result, let
\begin{equation}
s_i \in
\{
\texttt{absent},
\texttt{minor\_residue},
\texttt{major\_residue}
\}
\end{equation}
denote its source-residue label, where
\begin{itemize}
    \item \texttt{absent} indicates that neither the source concept nor any recognizable source-specific residue remains.
    \item \texttt{minor\_residue} indicates that a small but recognizable source-specific part, texture, color, or silhouette remains.
    \item \texttt{major\_residue} indicates that the source concept or substantial source-specific content remains clearly recognizable.
\end{itemize}

For the $i$-th edited result, let $r_i$ denote the fine-grained identity class returned by Qwen2.5-VL. We deterministically map
\texttt{TARGET\_ONLY} and \texttt{NEITHER} to
\texttt{absent}, \texttt{HYBRID\_TARGET\_DOMINANT} to
\texttt{minor\_residue}, and \texttt{OLD\_ONLY},
\texttt{HYBRID\_OLD\_DOMINANT}, and
\texttt{HYBRID\_BALANCED} to \texttt{major\_residue}.

Similarly, let
\begin{equation}
c_i \in
\{
\texttt{meaningful\_change},
\texttt{weak\_change},
\texttt{unchanged}
\}
\end{equation}
denote the regional-change label, where
\begin{itemize}
    \item \texttt{meaningful\_change} indicates a clear visual or semantic change within the annotated edit region.
    \item \texttt{weak\_change} indicates only limited or insufficient changes, such as slight movement, small texture variations, or minor color fluctuations.
    \item \texttt{unchanged} indicates that the annotated edit region remains essentially unchanged.
\end{itemize}

\paragraph{Source Residue Rate (SRR).}
SRR measures the proportion of edited results in which the source concept or recognizable source-specific content remains. Both minor and major residues are counted:
\begin{equation}
\mathrm{SRR}
=
\frac{1}{N}
\sum_{i=1}^{N}
\left\{
i \mid
s_i \in
\{
\texttt{minor\_residue},
\texttt{major\_residue}
\}
\right\}
\end{equation}
A lower SRR indicates more complete removal of the source concept.

\paragraph{Regional Edit Failure Rate (REFR).}
REFR measures the proportion of cases in which the annotated edit region undergoes only a weak change or remains nearly unchanged:
\begin{equation}
\mathrm{REFR}
=
\frac{1}{N}
\sum_{i=1}^{N}
\left\{
i \mid
c_i \in
\{
\mathrm{weak\_change},
\mathrm{unchanged}
\}
\right\}.
\end{equation}
A lower REFR indicates fewer insufficiently edited cases.

\paragraph{Effective Source-Erasure Rate (ESER).}
ESER counts a case as successful only when the source concept is no longer recognizable and the annotated edit region undergoes a meaningful change:
\begin{equation}
\mathrm{ESER}
=
1-\frac{1}{N}
\sum_{i=1}^{N}
\left\{
i \mid
s_i = \mathrm{absent}
\land
c_i = \mathrm{meaningful\_change}
\right\}.
\end{equation}
Unlike $1-\mathrm{SRR}$, ESER excludes cases in which the source concept is judged absent but the designated region remains nearly unchanged. It therefore provides a stricter measure of effective source removal.

\section{Automatic Source-Erasure Evaluation with Qwen2.5-VL}
\label{sec:supp_qwen_erasure}

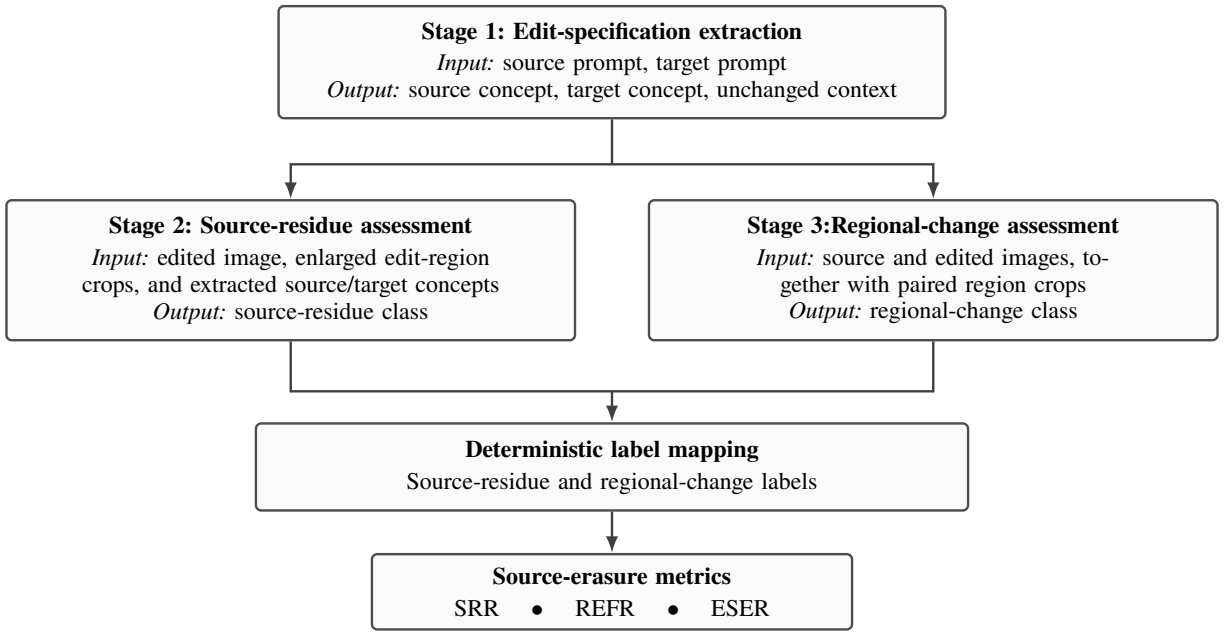
\begin{figure*}[t]
\centering
\begin{tikzpicture}[
    font=\small,
    >={Latex[length=2.2mm]},
    flowarrow/.style={
        ->,
        line width=0.9pt,
        draw=black!75
    },
    flowline/.style={
        line width=0.9pt,
        draw=black!75
    },
    mainbox/.style={
        draw=black!70,
        line width=0.9pt,
        rounded corners=2pt,
        fill=black!2,
        align=center,
        text width=8.4cm,
        minimum height=1.35cm,
        inner sep=6pt
    },
    branchbox/.style={
        draw=black!70,
        line width=0.9pt,
        rounded corners=2pt,
        fill=black!2,
        align=center,
        text width=7.1cm,
        minimum height=1.85cm,
        inner sep=6pt
    },
    mapbox/.style={
        draw=black!70,
        line width=0.9pt,
        rounded corners=2pt,
        fill=black!2,
        align=center,
        text width=9.0cm,
        minimum height=1.15cm,
        inner sep=6pt
    },
    metricbox/.style={
        draw=black!70,
        line width=0.9pt,
        rounded corners=2pt,
        fill=black!2,
        align=center,
        text width=6.0cm,
        minimum height=0.9cm,
        inner sep=5pt
    }
]

\node[mainbox] (spec) at (0,0) {
    \textbf{Stage 1: Edit-specification extraction}\\[2pt]
    \footnotesize
    \textit{Input:} source prompt, target prompt\\
    \textit{Output:} source concept, target concept,
    unchanged context
};

\node[branchbox] (residue) at (-4.25,-2.75) {
    \textbf{Stage 2: Source-residue assessment}\\[2pt]
    \footnotesize
    \textit{Input:} edited image, enlarged edit-region crops,
    and extracted source/target concepts\\
    \textit{Output:} source-residue class
};

\node[branchbox] (change) at (4.25,-2.75) {
    \textbf{Stage 3:Regional-change assessment}\\[2pt]
    \footnotesize
    \textit{Input:} source and edited images,
    together with paired region crops\\
    \textit{Output:} regional-change class
};

\node[mapbox] (mapping) at (0,-5.35) {
    \textbf{Deterministic label mapping}\\[2pt]
    Source-residue and regional-change labels
};

\node[metricbox] (metrics) at (0,-7.0) {
    \textbf{Source-erasure metrics}\\[2pt]
    SRR \quad $\bullet$ \quad REFR \quad $\bullet$ \quad ESER
};

\coordinate (split) at (0,-1.35);
\draw[flowline] (spec.south) -- (split);
\draw[flowarrow] (split) -| (residue.north);
\draw[flowarrow] (split) -| (change.north);

\coordinate (join) at (0,-4.35);
\draw[flowline] (residue.south) |- (join);
\draw[flowline] (change.south) |- (join);
\draw[flowarrow] (join) -- (mapping.north);

\draw[flowarrow] (mapping.south) -- (metrics.north);

\end{tikzpicture}

\caption{
Overview of the Qwen2.5-VL-based source-erasure evaluation. The evaluator first extracts the intended edit from the prompts. It then independently assesses source residue from the edited result and regional change from source--edited comparisons. The outputs are deterministically mapped to the labels used to compute SRR, REFR, and ESER.
}
\label{fig:qwen_source_erasure_pipeline}
\end{figure*}

We use Qwen2.5-VL-7B-Instruct as a frozen automatic evaluator for source-erasure analysis on replacement cases. As illustrated in Fig.~\ref{fig:qwen_source_erasure_pipeline}, the evaluator first extracts the source concept and target concept, and then predicts a class of source residue from the edited image and its enlarged region crops, which is deterministically mapped to one of three source-residue levels: \texttt{absent}, \texttt{minor\_residue}, or \texttt{major\_residue}. Separately, it compares the source and edited images to classify the regional change as \texttt{meaningful\_change}, \texttt{weak\_change}, or \texttt{unchanged}. These labels are used to compute SRR, REFR, and ESER. The complete evaluation prompts are provided in Sec.~\ref{sec:complete_qwen_prompts}.

\paragraph{Results.}
Table~\ref{tab:qwen_source_erasure} reports SRR, REFR, and ESER on the replacement cases of EditRegion-Bench and PIE-bench. Among methods without externally specified edit regions, PC-Edit achieves the lowest SRR and REFR and the highest ESER. Compared with the variant without $M_{\mathrm{src}}$, the full method substantially reduces SRR and REFR while improving ESER, confirming the effectiveness of the source-erasure footprint. These results indicate that the inversion-time source-erasure footprint more effectively suppresses recognizable source content while reducing insufficient edits.

\begin{table}[t]
\centering
\caption{
Qwen2.5-VL-based automatic source-erasure evaluation on the replacement cases of EditRegion-Bench and PIE-Bench.  Bold indicates the best result among methods without externally specified edit regions. $\dagger$ denotes methods provided with ground-truth edit boxes.
}
\label{tab:qwen_source_erasure}
\footnotesize
\setlength{\tabcolsep}{6pt}
\resizebox{\columnwidth}{!}{%
\begin{tabular}{lcccccc}
\toprule
\multirow[c]{2}{*}[-1ex]{Method}&
\multicolumn{3}{c}{\textbf{EditRegion-Bench}}
&
\multicolumn{3}{c}{\textbf{PIE-Bench}}
\\
\cmidrule(lr){2-4}
\cmidrule(lr){5-7}
& SRR$\downarrow$
& REFR$\downarrow$
& ESER$\uparrow$
& SRR$\downarrow$
& REFR$\downarrow$
& ESER$\uparrow$
\\
\midrule
PnPInversion
& 32.66 & 22.58 & 62.10
& 37.50 & 47.50 & 45.00
\\
MasaCtrl
& 65.32 & 50.40 & 27.42
& 46.25 & 58.75 & 30.00
\\
RF-Inversion
& 31.45 & 17.34 & 65.73
& 38.75 & 15.00 & 58.75
\\
Stable Flow
& 63.71 & 55.65 & 27.02
& 55.00 & 67.50 & 22.50
\\
FlowEdit
& 24.60 & 19.35 & 70.56
& 37.50 & 32.50 & 55.00
\\
FYS
& 21.37 & 18.95 & 73.39
& 31.25 & 25.00 & 60.00
\\
RF-Solver-Edit
& 20.56 & 20.16 & 73.39
& 28.75 & 17.50 & 62.50
\\
\midrule
FLUX.1-Fill$^{\dagger}$
& 14.71 & 4.23 & 81.68
& 27.50 & 21.25 & 66.25
\\
KV-Edit$^{\dagger}$
& 29.84 & 22.98 & 65.32
& 37.50 & 38.75 & 50.00
\\
\midrule
PC-Edit w/o $M_{\mathrm{src}}$
& 23.39 & 14.52 & 72.18
& 27.50 & 27.50 & 58.75
\\
\textbf{PC-Edit (ours)}
& \textbf{15.53}
& \textbf{11.50}
& \textbf{82.44}
& \textbf{21.25} & \textbf{23.75} & \textbf{67.50}
\\
\bottomrule
\end{tabular}
}
\end{table}

\section{More Qualitative Results}
\label{sec:more-qualitative}

\begin{figure*}[t]
\centering
\includegraphics[width=\textwidth]{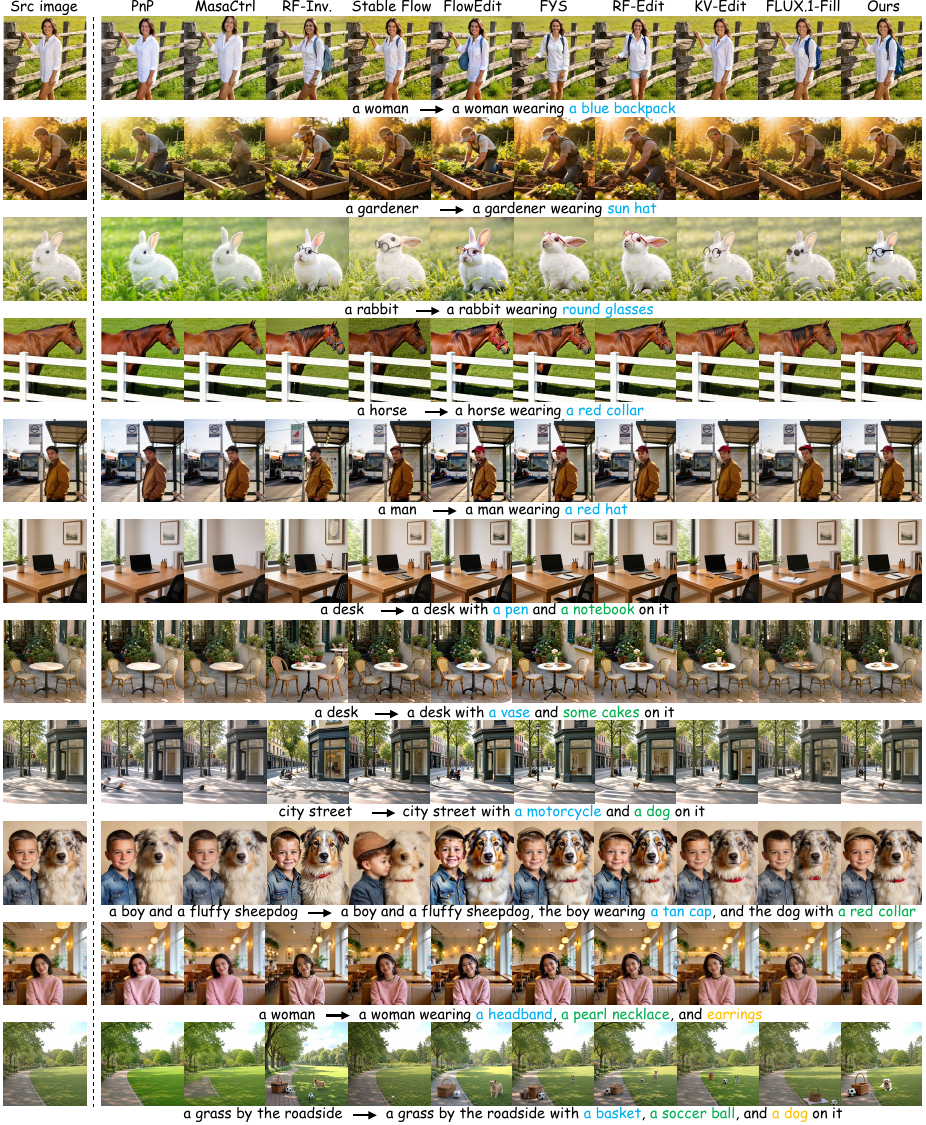}
\caption{Additional qualitative comparisons on object addition. PC-Edit introduces the requested objects while better preserving unrelated content.}
\label{fig:reserved-addition}
\end{figure*}

\begin{figure*}[t]
\centering
\includegraphics[width=\textwidth]{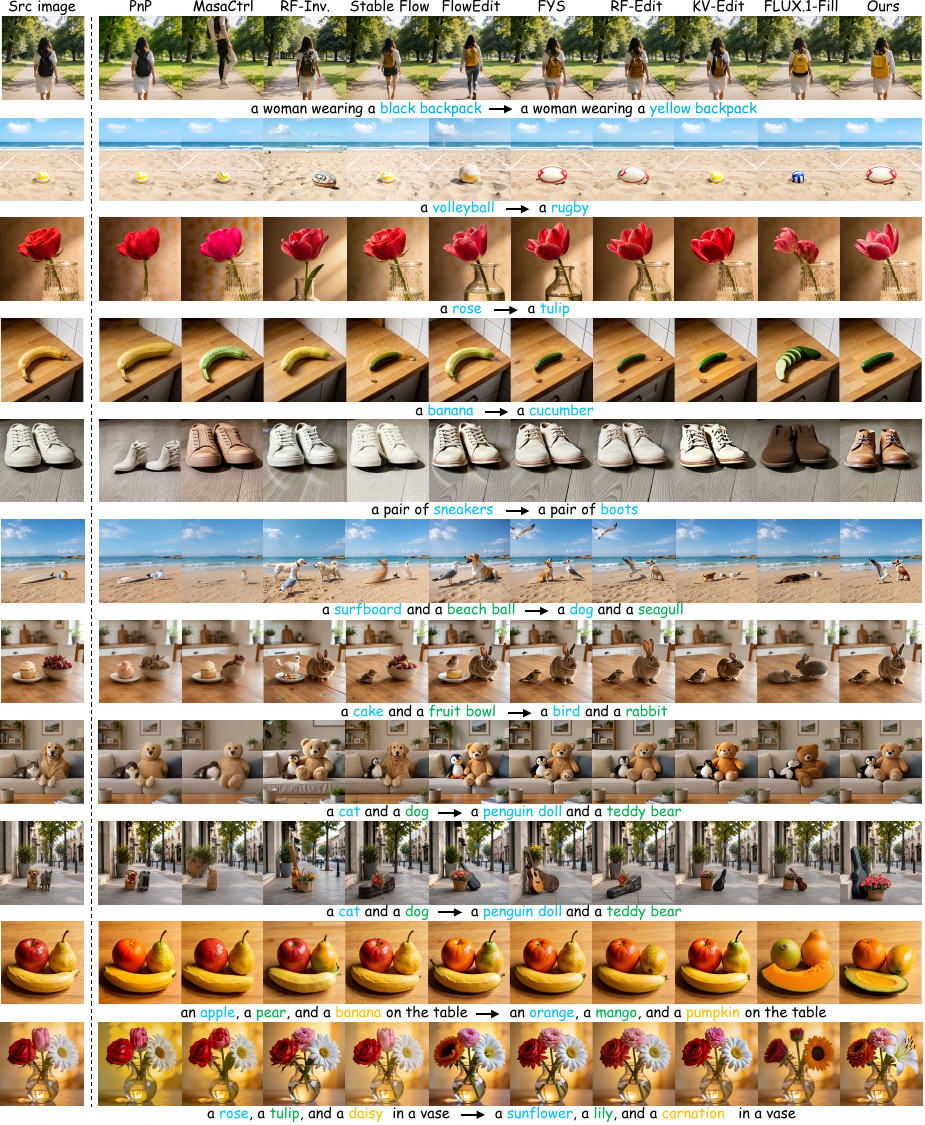}
\caption{Additional qualitative comparisons on object replacement. PC-Edit more completely removes the source objects and generates the target objects while better preserving unrelated content.}
\label{fig:reserved-replacement}
\end{figure*}

Figures~\ref{fig:reserved-addition} and~\ref{fig:reserved-replacement} provide additional qualitative comparisons on object addition and replacement, respectively, complementing the results reported in the main paper. For addition, PC-Edit more consistently introduces the requested objects while preserving unrelated scene content. For replacement, it better removes the source objects, accommodates the shapes of the target objects, and limits unintended changes outside the edited regions. These examples further demonstrate the effectiveness of prompt-contrastive region discovery and region-guided source-feature injection across diverse editing scenarios.

\clearpage
\onecolumn

\section{Complete Qwen2.5-VL Evaluation Prompts}
\label{sec:complete_qwen_prompts}

The following prompts are reproduced exactly for completeness. Placeholders enclosed by braces are replaced with the corresponding case-specific information during evaluation.

\subsection{Stage A: Edit-Specification Extraction}
\label{sec:qwen_prompt_stage_a}

\noindent\textbf{System prompt}

\begin{PromptBlock}
You precisely compare captions. Return JSON only.
\end{PromptBlock}

\noindent\textbf{User prompt}

\begin{PromptBlock}
Extract only the intended visual edit from two captions.
Do not inspect any image. Separate changed concepts from shared context.

SOURCE PROMPT: {src}
TARGET PROMPT: {dst}
TASK TYPE: {edit_type}

For category replacement, old_concept is the complete old object identity and new_concept is the complete new object identity.

For attribute replacement, old_concept/new_concept are only the changed attributes, while the shared object belongs in shared_context.

For addition, old_concept is "none".

Return JSON only:
{
  "old_concept": "specific identity/attribute or none",
  "new_concept": "specific identity/attribute",
  "shared_context": ["unchanged concepts"],
  "edit_subtype":
    "category_replacement |
     attribute_replacement |
     addition |
     multi_edit"
}
\end{PromptBlock}

\subsection{Stage B: Source Residue Classification}
\label{sec:qwen_prompt_stage_b}

For this stage, the evaluator receives the edited full image and the
enlarged crops of the annotated edit regions. The source image is not
provided.

\noindent\textbf{System prompt}

\begin{PromptBlock}
Judge literal visible evidence in the edited image only.
Return JSON only.
\end{PromptBlock}

\noindent\textbf{User prompt}

\begin{PromptBlock}
Inspect only the EDITED result, especially each numbered region.

OLD CONCEPT THAT SHOULD HAVE BEEN REMOVED:
{old_concept}

REQUESTED TARGET CONCEPT:
{new_concept}

First describe the visible identity, colors, texture/material, diagnostic parts, and silhouette.

Then choose exactly one mutually-exclusive class:

- OLD_ONLY:
  The old concept remains recognizable and the target is absent.

- TARGET_ONLY:
  The target is recognizable and no old-specific cue remains.

- HYBRID_OLD_DOMINANT:
  Both old and target concepts contribute, with strong or extensive old identity remaining.

- HYBRID_TARGET_DOMINANT:
  The target is recognizable, but a small old-specific part, color, texture, outline, or silhouette remains recognizable.

- HYBRID_BALANCED:
  Both old and target identities are substantially visible.

- NEITHER:
  Neither the old nor the target concept is recognizable.

For category replacement, explicitly check old-specific body shape, silhouette, material/texture, and diagnostic parts even when the face or surface resembles the target.

For attribute replacement, any recognizable old color or texture is old residue. The shared object category is not residue. Common features shared by the old and target concepts do not count as residue.

Do not use a source image and do not reward the requested text.

Return exactly one JSON object:
{
  "visible_description":
    "literal visible identity and old/target cues",
  "class":
    "OLD_ONLY |
     TARGET_ONLY |
     HYBRID_OLD_DOMINANT |
     HYBRID_TARGET_DOMINANT |
     HYBRID_BALANCED |
     NEITHER",
  "evidence":
    "short decisive visible evidence"
}
\end{PromptBlock}

\subsection{Stage C: Regional-Change Classification}
\label{sec:qwen_prompt_stage_c}

For this stage, the evaluator receives the source and edited full
images together with paired source and edited crops of the annotated
regions.

\noindent\textbf{System prompt}

\begin{PromptBlock}
Judge literal visible evidence in the edited image only.
Return JSON only.
\end{PromptBlock}

\noindent\textbf{User prompt}

\begin{PromptBlock}
Compare SOURCE versus EDITED only inside and immediately around the numbered regions.

Choose exactly one class:

- MEANINGFUL_CHANGE:
  A clear object or attribute rewrite is visible.

- WEAK_CHANGE:
  Only movement, tiny texture or color fluctuation, or reconstruction noise is visible.

- UNCHANGED:
  Essentially no visible change is present.

This question is only about change magnitude, not target success.

Return exactly:
{
  "class":
    "MEANINGFUL_CHANGE |
     WEAK_CHANGE |
     UNCHANGED",
  "evidence":
    "literal visible difference"
}
\end{PromptBlock}


\end{document}